# Tip Manipulation in Soft Everting Robots via Wall Retraction and Deployable Fingers

Nelson Badillo Pérez [1,†], Niccolò Pagliarani [2,†], Matteo Cianchetti [2] and Robert D. Howe [1]

1 John A. Paulson School of Engineering and Applied Sciences, Harvard University, Cambridge, MA, USA

2 The BioRobotics Institute, Scuola Superiore Sant'Anna, Pisa, Italy

† These authors contributed equally to this work.
Corresponding author: Robert D. Howe, howe@seas.harvard.edu

## ABSTRACT

Soft everting robots can traverse long, confined paths by continuously growing, yet active interaction remains limited to a single tool fixed at or near the tip, unable to be repositioned on demand and difficult to reconcile with the robot's soft body. We introduce a tip-manipulation and multi-tool deployment strategy for soft everting robots based on wall retraction, implemented with a base roller assembly that independently meters membrane flow in the outer wall while a tail spool regulates growth in the internal tail section. Coordinated wall and tail actuation decouples robot length from membrane-material position, enabling membrane-mounted devices to be transported, exposed, and repositioned at selected locations near the distal tip. We pair this capability with ultralight pleated inflatable fingers integrated into the membrane, fabricated from TPU-coated nylon with an internal airtight bladder. The fingers achieve large bending at low pressures (approximately 100° at 50 kPa in high-pleat designs) and generate blocking forces up to ~1.9 N, while remaining limp during transport. The system demonstrates adaptive grasping across diverse household objects (≈21–550 g; ≈16–200 mm), three-dimensional object manipulation and stacking, environmentally braced extension, distal camera panning for confined-space inspection, and controlled sequential payload delivery. These results enable embodied and reversible tip manipulation for soft growing robots in cluttered and tortuous environments.

## SUMMARY

A soft everting robot that drives its wall and tail independently can deploy wall-mounted fingers for grasping at any location. Adding wall rollers to vine robots decouples membrane motion and robot length, so wall-mounted fingers can be deployed anywhere for tip grasping.

## INTRODUCTION

Vine robots have an unequaled ability to reach across long distances but have limited capability for dexterous interactions. These soft everting robots (SERs) navigate by extending new material at the tip (*1*–*3*). Research to date has focused predominantly on locomotion, with SERs serving as carriers for cameras, sensors, and rigid end-effectors (*1*, *4*, *5*, *10*). This enables applications such as inspection, search-and-rescue operations, and minimally invasive medical procedures (*6*–*10*). The ability to actively manipulate objects, however, remains an unsolved challenge, resulting in largely passive tip interactions. A major constraint is that the eversion process entails large deformations of the SER body membrane, which prevents the attachment of sizeable structures (*4*). In addition, effectors that are attached to the robot body are deployed at predetermined locations as the robot extends. Representative approaches to distal manipulation in soft everting robots can be grouped into three main strategies. The first relies on end-effector platforms that remain at the everting front despite continuous membrane turnover (*4*). Although these systems enable distal sensing and tool use, their added mass and stiffness can reduce mobility and compromise tip compliance. A second strategy exploits the everting body itself to engulf or disgorge objects at the tip (*11*,

*12*). This preserves the compliant and lightweight nature of the robot but limits use cases to mostly small objects or situations where an external insertion force can be applied. Magnetically actuated distal modules offer a third approach, enabling remote steering, force transmission, and tool interaction near the tip (*13*), but their dependence on external magnetic-field generators restricts their use to specialized environments. Collectively, these studies demonstrate the growing interest in manipulation for soft everting robots, while highlighting the lack of a general solution that preserves the advantages of tip growth.

Here, we endow SERs with controllable tip manipulation and tool placement based on wall retraction (Fig. 1). A set of rollers at the base of the robot can extend and retract the outer wall. Coordinating the motion of these wall actuators with the motion of the tail actuator decouples body length and membrane position, so components mounted on the membrane can be relocated anywhere along the body, in particular at the tip for grasping. In addition, by actuating the wall rollers on opposite sides of the body at different rates, the robot can be steered across a range of angles in three dimensions.

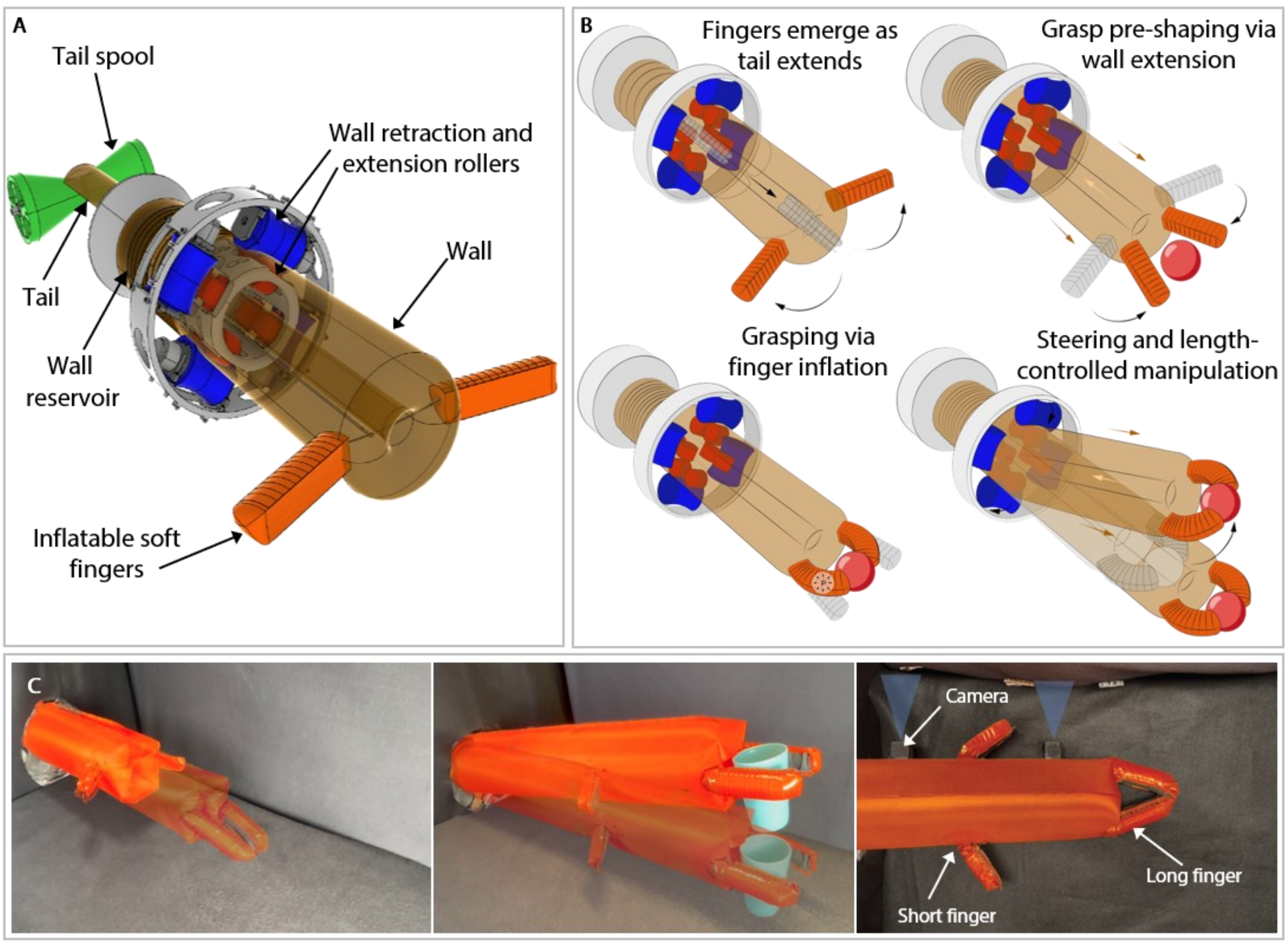

**Fig. 1. Architecture for tip manipulation in soft everting robots. (A)** The robot consists of a membrane that extends from a tail spool, through the inside of the soft pressure-driven body to the everting tip, and back along the outer wall to a set of four rollers. Motors on the tail spool and wall rollers can move the membrane to independently control robot length and the position of inflatable fingers mounted on the membrane. **(B)** Tip manipulation sequence: fingers are transported during growth, exposed at the distal tip, positioned relative to the target object by wall extension/retraction, and inflated to grasp the object. The object may then be manipulated through robot body steering and length control. **(C)** Representative capabilities, including regulated growth, object manipulation, and deployment of multiple tools.

We also present novel ultralightweight inflatable fingers that mount on the SER membrane as representative soft appendages that integrate with the robot body while generating controllable kinematic constraints and contact forces on grasped objects. Using a pleated structure sewn from flat fabric, the fingers bend under low inflation pressures to provide variable gripping force and grasp stiffness. By coordinating their inflation with their position along the robot body via wall and tail actuation, these fingers are capable of grasping a wide range of object sizes and shapes, while adding negligible weight and preserving SER compliance.

In the following sections, we detail the design and implementation of the wall retraction mechanism and derive the kinematic relationships between actuation and tip and wall motion. We then present the design and fabrication of the inflatable fingers and characterize their performance on the bench and in grasping experiments. Next, we demonstrate how the same architecture enables grasping of everyday objects spanning 21–550 g in weight and 16–200 mm in characteristic dimension, three-dimensional object manipulation and stacking, environmental bracing to extend unsupported reach, deployable camera panning for confined-space inspection, sequential deployment of multiple tools, and controlled payload delivery. The paper concludes by discussing extensions toward closed-loop sensing, distributed tool deployment, and a variety of applications.

## RESULTS

### Wall retraction in soft everting robots

As shown in Fig. 1A, the robot consists of a soft everting body. At the base of the robot, a wall retraction device controls how the membrane moves using a set of four roller assemblies. Each actuated roller (shown in blue in Fig. 1) engages two passive rollers (shown in red) to provide traction against the membrane. This allows the wall to be advanced or retracted while the tail is moved independently, thereby decoupling robot length from membrane-material position and enabling membrane-mounted components to be repositioned on demand (movie S1). Membrane-mounted tools can be stored within the tail and transported to the distal tip without permanently occupying the end-effector region. Movie S2 demonstrates the sequential deployment and actuation of two cameras and two ultra-light inflatable finger sets, followed by their retraction through reversal of the membrane flow. These fingers are made of thin, compliant fabric and remain limp until they are inflated. In their unpressurized state, they add negligible stiffness or bulk. When they are everted at the tip and inflated, however, the fingers bend smoothly and conform to objects, enabling adaptive grasping.

An example of a grasping process is illustrated in Fig. 1B. As the robot grows by unrolling material from the tail spool, the inflatable fingers emerge at the tip. Before contacting an object, the robot can pre-shape the grasp by moving the outer wall to adjust the aperture and orientation of the fingers demonstrated in movie S3. Once positioned, inflating the fingers causes them to bend and contact the object. The robot can then lift, translate, and reorient the grasped object in three dimensions by coordinating the tail-spool and wall-roller motors to regulate body length and steering. This coordinated motion enables precise sequential manipulation tasks, such as placing and stacking one object on top of another, as demonstrated in movie S3. Finally, deflating the fingers or retracting the wall releases the object at the desired location. Wall retraction adds capabilities while preserving the main benefits of SERs. As noted above, wall motion decouples body length and membrane position, so tools and sensors attached to the membrane can be positioned anywhere along the length of the robot's outer wall. Using the rollers to advance the membrane on one side of the body while keeping the other side fixed results in deflection of the robot body, so it can be steered across a range of angles in two directions. The robot body remains soft, which enables compliant grasping and low-force interactions. It can squeeze through small openings and confined spaces and is undamaged by large deformations of the robot body.

Besides the addition of four wall actuators, an operational difference from conventional SERs is that wall retraction produces sliding motion at contacts between the outer wall of the robot and the surrounding environment. The resulting frictional forces can limit wall motion when there are extended contact areas or when the robot pushes through confined passages. The robot can, however, function as a traditional everting vine robot under these conditions.

## Motion modes

The membrane actuation strategy creates five modes for independent locomotion and positioning of membrane-mounted tools through coordinated control of the tail spool, which regulates material supplied to the inner tail, and the wall rollers, which control outer membrane motion (Fig. 2 and movie S1). In conventional vine robot tail *eversion growth* (Fig. 2, column i), internal pressure and motorized tail unspooling advance the distal tip while the outer wall remains stationary. This mode allows the robot to extend through long or confined paths. In *wall retraction* (column ii), the wall rollers retract outer membrane material toward the base while the tail spool remains fixed. This shortens the deployed body and causes the tip to retreat via wall-induced inversion. In membrane *cycling* (column iii), the tail unspools while the wall rollers retract material at the same rate. As a result, membrane material is transported through the robot while the distal tip position and body length remain approximately fixed. This mode is central for tool deployment, because it can transport membrane-mounted fingers or sensors across the robot body without moving the robot pose.

In *body extension* (column iv), the wall and tail are advanced at the same rate, lengthening the body while keeping membrane-mounted components fixed at the distal tip. This allows the robot to reposition a deployed tool without changing its location on the tip, thus preserving grasp configurations during manipulation. Finally, *steering* (column v) is produced by asymmetric wall motion: retracting or extending one side of the wall while holding the opposite side fixed causes the body to bend and changes the tip angle. The four actuated wall rollers allow the robot to steer in pitch and yaw.

Experiments in Fig. 2B demonstrate these five modes by tracking the motion of the tip, wall, tail, and tip angle. These modes can be sequenced to produce complex behaviors. In the grasping process in Fig. 1B the robot first approaches a target by *eversion growth*, then uses *wall retraction* and *cycling* to bring the inflatable fingers to the desired position at the tip, inflates the fingers to grasp the object, and finally uses *body extension* and *steering* to reposition the grasped object while maintaining the grasp. These modes may be blended by regulating actuation rates in conjunction with length changes.

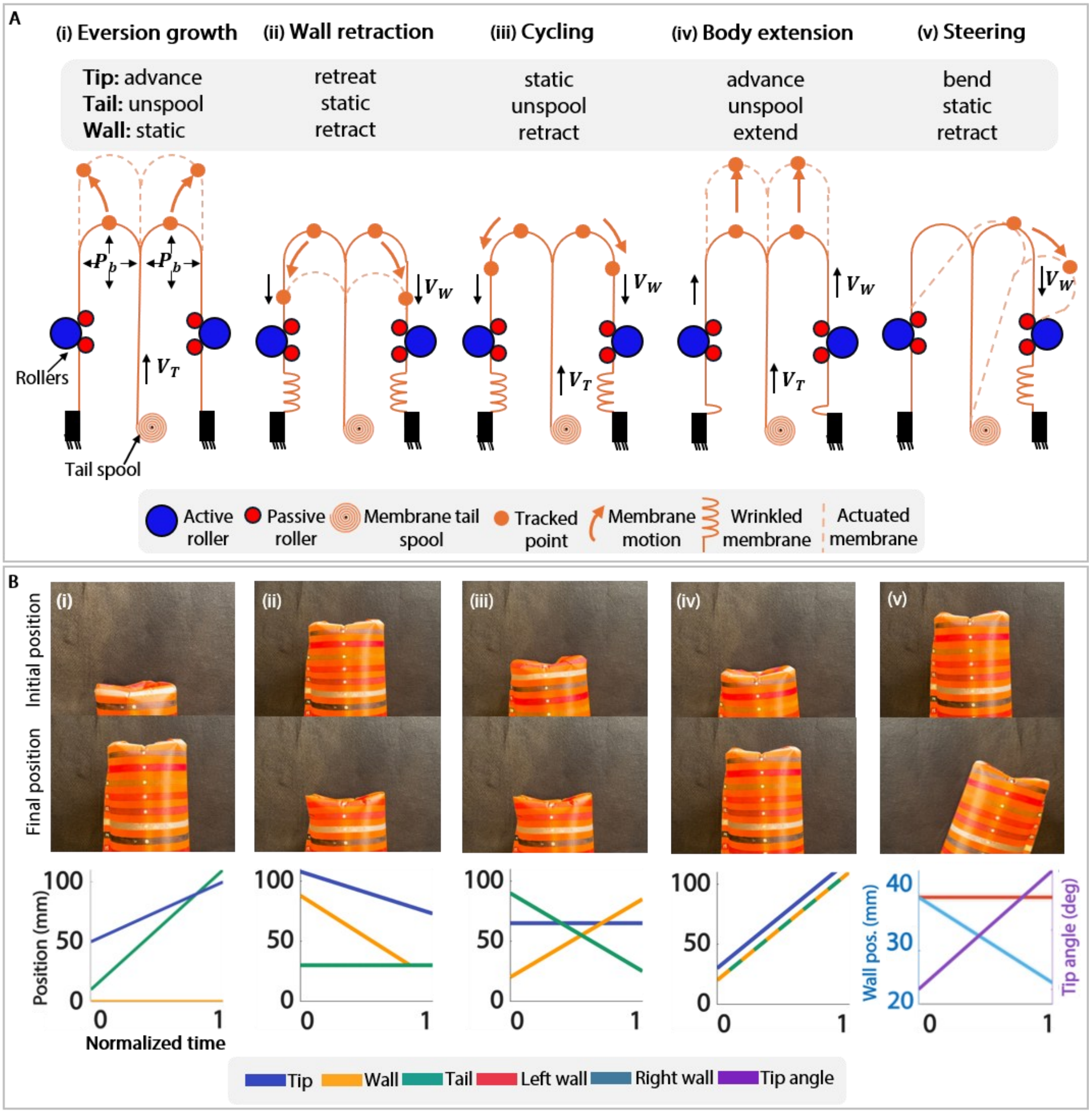

**Fig. 2. Coordinated wall–tail actuation generates programmable membrane-flow modes. (A)** Five motion modes produced by coordinated control of the tail spool and wall rollers: eversion growth, wall retraction, membrane cycling, body extension, and steering. Active rollers are shown in blue, passive rollers in red, the membrane tail spool in an orange spiral pattern, and tracked membrane points in orange. **(B)** Experimental demonstration of each mode, showing initial and final robot configurations. Curves report the normalized changes in tip, wall, tail, left-wall, right-wall, and tip-angle states over each commanded motion.

## Kinematic control of tip position and membrane transport

Implementation of the motion modes requires the inverse kinematics solution. We formulate this solution in terms of three spaces (Fig. 3): the actuator space, the body length space, and the tip space. Actuator space is defined by the tail spool and wall roller encoder position, $\mathbf{q} \in \mathbb{R}^5$. The tip space is defined by the position of the center of the tip $\boldsymbol{P_E} \in \mathbb{R}^3$and the membrane position $S_M \in \mathbb{R}^1$. The latter is a scalar state variable that is measured relative to the distal tip point $S_o$, (Fig. 3A).

For *body length space*, the robot is modeled as three idealized sections: a cylindrical outer wall extending from the "base platform" plane containing the wall rollers; a doubly-curved half-torus tip smoothly connected to the wall cylinder at the "end-effector platform" plane, forming the distal everting surface; and a linear tail extending back from point *G* at the torus center to the tail spool.

To relate actuator and body length spaces, we describe "virtual strings" extending from the four wall rollers $M_{Wi}$ axially along the cylindrical wall, around the half-torus, and back along the tail spool, with segment lengths denoted $L_{Wi}$, $L_{Torus}$ , and $L_{Tail}$, respectively (Fig. 3A(i)). We assume $L_{Torus}$ is constant and corresponds to a semicircular fold of radius $R_{Body}/2$, giving an arc length of $(\pi/2)R_{Body}$, where $R_{Body}$ is the membrane cylinder radius. The tip center $\mathbf{P}_E$ lies $R_{Body}/2$ axially outward from G.

To relate the tip position $\boldsymbol{P}_E$ to the actuator positions $\boldsymbol{q}$, the system is modeled as a parallel robot analogous to a 3-DOF Stewart Platform (*14*). The wall roller motors control the lengths of the virtual strings $L_{Wi}$ which function as "prismatic linkages," and the tail spool motor controls the length of the tail $L_{Tail}$. These actuators connect the "base platform" and the "end-effector platform" defined above.

A detailed kinematic solution is presented in Supplementary Section S5. The closed-chain inverse kinematics are obtained from the loop-closure equations (Eqs. S7 to S13), which map the desired tip-space state ($\boldsymbol{P_E}$, $S_M$) to the five actuator positions. Although the robot has five actuators, its tip space has four independent variables: the three components of $\boldsymbol{P_E}$ and $S_M$. The end-effector orientation is uniquely determined by $\boldsymbol{P_E}$ through the kinematic constraint in (*15*). Moreover, the four wall-string lengths are geometrically coupled because they connect to the same base and end-effector planes. Their coordination with the tail length controls $\boldsymbol{P_E}$ and $S_M$.

For the membrane position $S_M$, an arbitrary tagged point is parameterized through the tail, torus, and a selected wall string as a one-dimensional curve (Fig. 3A(ii,iii)). Wall and tail length changes can drive two motions at once: alter robot shape, repositioning the distal endpoint $\boldsymbol{P}_E$, and a drift of $S_M$, along the path. Because $\boldsymbol{P}_E$ determines virtual membrane length changes while $S_M$ depends on the asymmetry between tail and average wall changes, the two effects can be handled in sequence: first calculating the virtual strings that correspond to the desired $\boldsymbol{P}_E$, then calculating the drift $\Delta S_M$ to deliver the membrane-mounted tool to its target location on the body. Under this approach, the drift $\Delta S_M$ takes a simple form,

$$\Delta S_M = \frac{1}{2}\left(\Delta L_{Tail} - \Delta L_{W\ Avg}\right) \qquad (1)$$

where $\Delta L_{Tail}$ is the change in tail length and $\Delta L_{W\ Avg}$ is the change in the average length of the wall strings. Equal changes in tail and average wall length leave the tool in place relative to the tip, while asymmetry between them drives eversion or inversion and transports the tool toward or away from the tip.

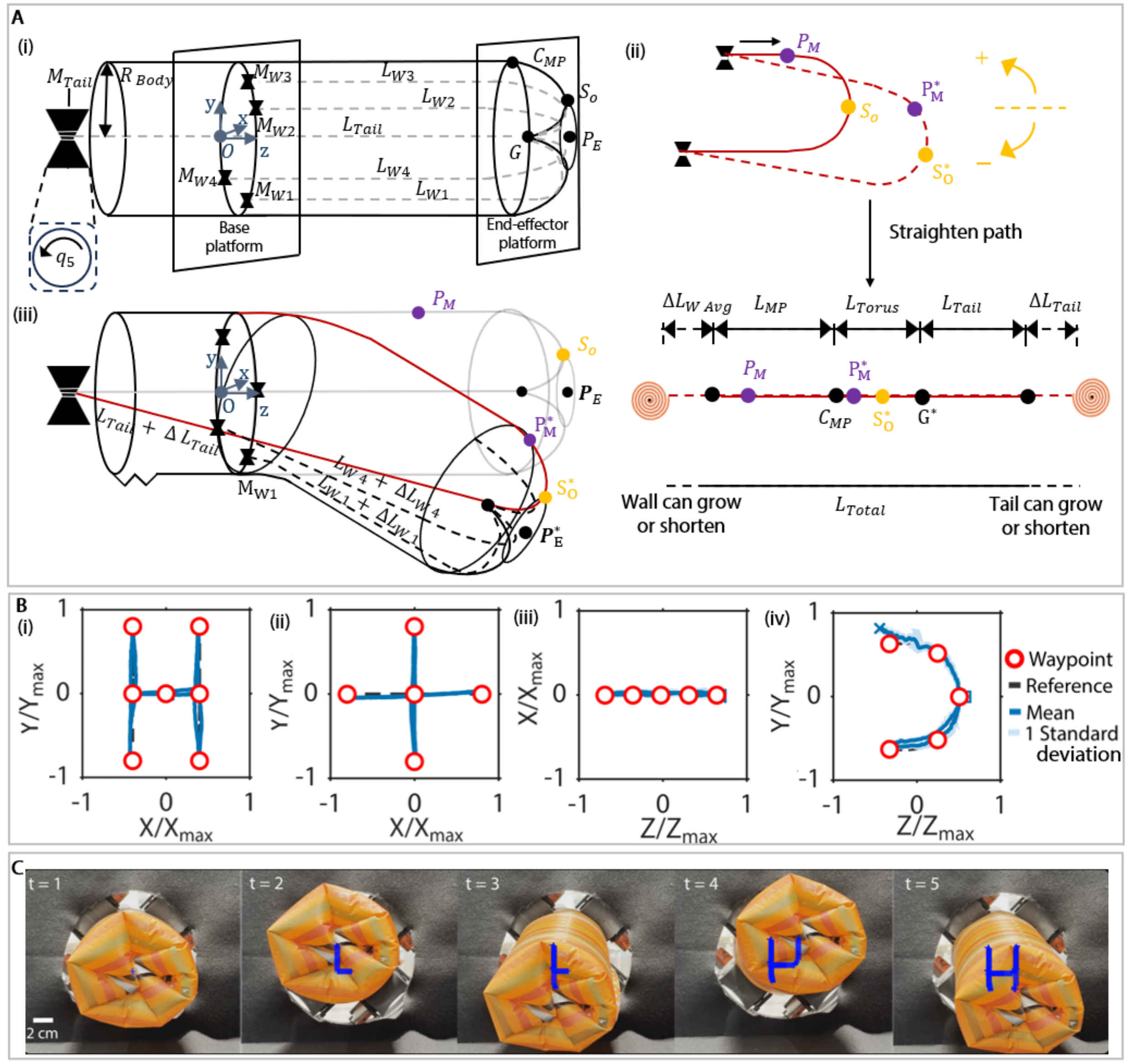

**Fig. 3. Kinematic model and open-loop control of distal pose and membrane transport. (A)** Kinematic representation of the soft everting robot. **(i)** The system is modeled as a closed-chain parallel mechanism in which the tail spool and four wall rollers regulate tail length $L_{Tail}$ and wall-string lengths $L_{Wi}$, thereby determining the distal position $\mathbf{P}_E$ and its desired distal position $P_E^*$ **(ii)** The position $P_M$ of a tagged membrane point and its desired position $P_M^*$ is parameterized along a one-dimensional path comprising the tail, toroidal tip, and selected wall segment; the corresponding straightened path illustrates the total membrane length $L_{Total}$, distal reference position $S_O$ and desired reference $S_O^*$ **(iii)** Coordinated changes in tail and wall lengths modify the robot configuration while transporting the tagged membrane point. **(B)** Open-loop validation for **(i)** H-shaped and **(ii)** '+'-shaped planar positioning, **(iii)** axial body extension and retraction, and **(iv)** membrane cycling. Red circles indicate commanded waypoints, black dashed lines the reference trajectories, and blue curves mean measured trajectories over five trials; shaded regions denote ±1 SD. Coordinates are normalized by the maximum commanded displacement. **(C)** Representative sequence of the H-shaped trajectory. Blue lines indicate the traced path.

We validated the kinematics framework using open-loop execution of representative trajectories that combine the five motion modes. Waypoints were commanded while motion was tracked with a lightweight electromagnetic 6-DOF sensor attached to the membrane, either at the distal cap center to measure $\mathbf{P}_E$ or at the distal toroidal fold to measure $\mathbf{P}_M$. Figure 3B(i–iv) shows four normalized trajectories: H and "+"- shaped planar positioning, axial extension and retraction, and membrane cycling with maximum excursions of 40, 80, 200, and 60 mm, respectively. Mean waypoint errors were 5.0 to 6.8% of the maximum excursion (2.7, 5.1, 10.0, and 3.4 mm for the H, +, axial extension, and cycling trajectories, respectively; $n$ = 5 trials each). Residual errors arise from passive material drift caused by asymmetric roller traction and from unmodeled construction tolerances. These results confirm accurate tip and membrane motion within the quasi-static, pre-buckling regime.

## Stowable pleated fingers

Wall and tail actuators can transport membrane-mounted tools to arbitrary locations on the robot body, but these tools must not hinder eversion at the tip. We developed inflatable fingers based on a pleated fabric architecture (Fig. 4A) as a membrane-mounted tool because they collapse into a thin, compliant state during transport within the tail and provide controllable contact forces and shape adaptation for grasping after deployment. Each finger consists of a smooth bottom layer, a pleated top layer, and an internal airtight TPU bladder. The fingertip and bottom layer provide the main contact surface during grasping, while the pleated top layer enables pressure-driven shape change. In the unpressurized state, the finger remains flat and compliant for transport. When pressurized, the bladder expands and the finger first extends in a straight cylindrical pose, then with increasing pressure the pleats slide and unfold, producing bending (Fig. 4A,B). Releasing pressure deflates the bladder, closes the pleats, and moves the finger back toward a straight configuration. The complete fabrication workflow is shown in Fig. S6.

Mechanically, finger bending results from an interplay of a pressure–driven bending moment, sliding of the pleats in the top layer, and the springlike behavior of the bottom layer. When initially pressurized, the finger inflates to its full diameter, set by the circumference of the inextensible membrane material. As pressure increases, there is a growing outward force on the distal end. In response, the pleats begin to slide and unfold, increasing the length of the top. The bottom layer is composed of a single layer of inextensible membrane material, so it bends as the top layer extends. The bottom layer is shaped as a curvature-stiffened strip spring (similar to a measuring tape blade) which bends in response to the pressure-generated bending moment. When pressure decreases, the bending moment diminishes and the bottom spring straightens. This generates a compressive force on the pleats. Because the bladder is forced against the pleats by pressure, the pleats slide and refold rather than buckle. This bending process is highly repeatable, with hundreds of cycles observed.

To characterize the design space, we measured the effect of pleat number on pressure-driven bending (Fig. 4C) for a fixed finger length. Across all designs, the bending angle increased monotonically with pressure from 0 to 50 kPa. Increasing the number of pleats produced larger bending angles at the same pressure, with the highest-pleat design ($N_p$ = 16) reaching approximately 100° at 50 kPa. Qualitatively, this trend can be explained by the role of the pleated dorsal structure: adding pleats reduces the effective bending resistance of the top layer by distributing opening and sliding over more fold units, allowing the finger to accommodate greater curvature under the same pressure.

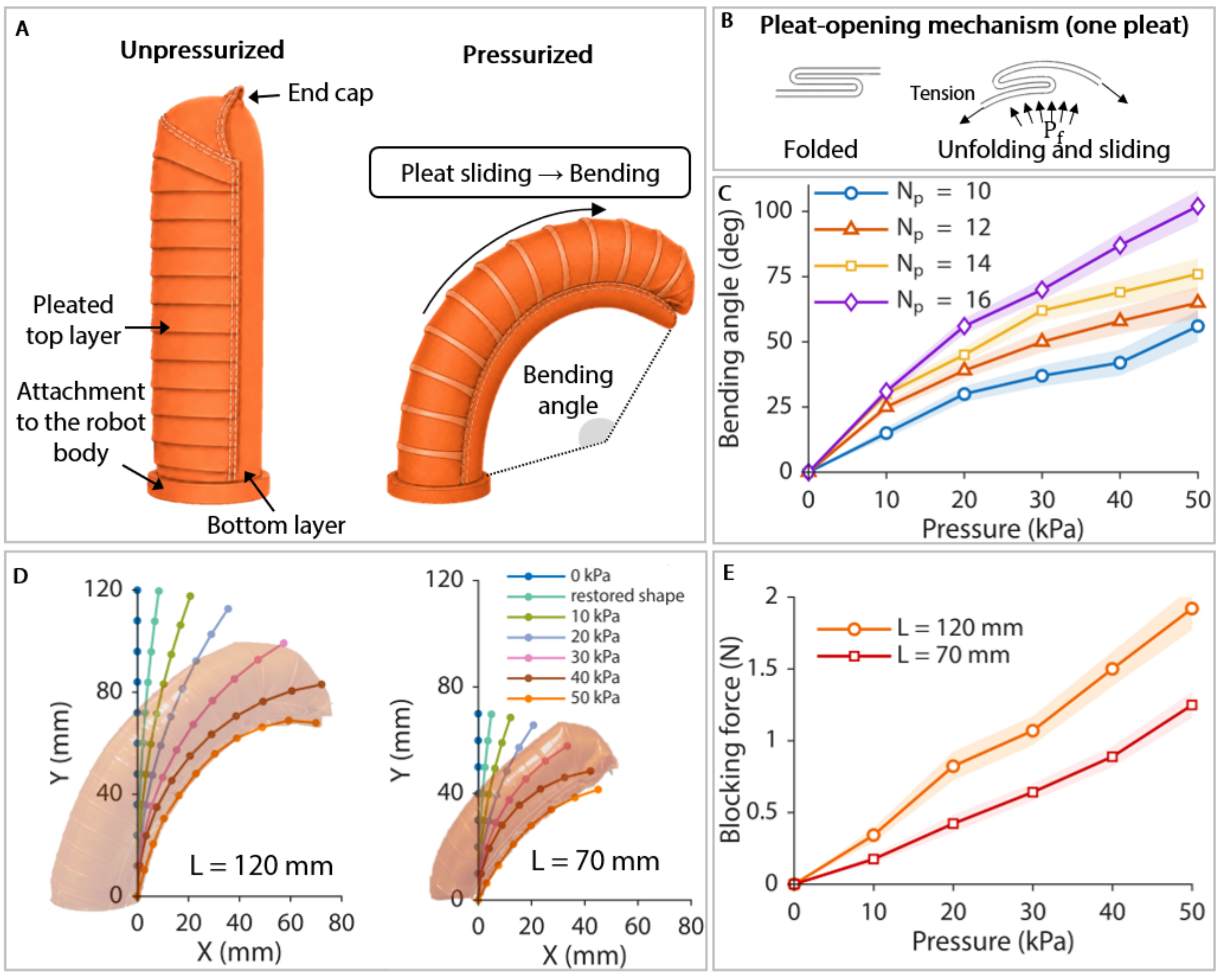

**Fig. 4. Design and characterization of stowable pleated inflatable fingers. (A)** Finger architecture in the unpressurized and pressurized states, showing the pleated top layer, bottom layer, end cap, and attachment to the robot body. Inflation induces pleat unfolding and sliding, lengthening the top layer and producing bending. **(B)** Pressure-driven opening of a single pleat, showing its folded state and unfolding and sliding under the pressure $P_f$. **(C)** Mean bending angle versus pressure for fingers with $N_p = 10$, 12, 14, and 16 pleats. **(D)** Pressure-dependent centerline profiles for fingers with lengths $L = 120$ and $L = 70$ mm, including the initial configuration at 0 kPa and the restored shape after depressurization. **(E)** Mean blocking force versus pressure for both finger lengths. In **(C)** and **(E)**, symbols indicate mean values, lines connect pressure conditions, and shaded regions denote ±1 standard deviation across five trials per condition.

We also evaluated how finger length affects shape evolution and force output. Fig. 4D shows the measured bottom contact surface shapes of two fingers with 12 pleats, with lengths of $L = 70$ and 120 mm, as pressure increased. In both cases, pressure provided continuous control over curvature, while the longer finger generated a larger swept grasping envelope. Finally, we measured the blocking force (force applied to a fixed surface; see Materials and Methods below for details) as a function of pressure (Fig. 4E). Blocking force increased monotonically for both lengths, reaching approximately 1.9 N at 50 kPa for the 120-mm finger and approximately 1.25 N for the 70-mm finger. Together, these results show that the pleated fingers provide a deployable grasping interface whose curvature and force can be set through design geometry and operating pressure, while remaining compatible with transport inside the soft everting body. Integration of the finger module into the everting membrane body is shown in Fig. S7.

## Body–finger coupling in grasping

In addition to the finger properties characterized above, the grasping behavior of the robot depends on the geometry and compliance of the everting body on which the fingers are mounted. The fingers can be positioned at any angle $\theta_{toroid}$ on the distal torus section through eversion and inversion of the body, which sets the maximum aperture of the grasp (Fig. 5A). The robot can then grasp objects by curling the fingers through inflation and/or by moving the finger base to a smaller $\theta_{toroid}$ (Fig. 5B).

Each finger integrates into the TPU-coated nylon robot body through a heat-sealed circular interface, providing an airtight connection without a rigid frame that would hinder eversion. Two monofilament anchors span the finger base as opening limiters, preventing transverse spread under load-induced buckling while still permitting the finger to collapse and travel during eversion.

To understand how finger inflation and toroid position interact to produce useful grasps, we analyze the body–finger coupling in terms of two regimes, enclosure and fingertip grasps. Figure 5A defines the two configuration variables: the finger bending angle, $\theta_{bending}$, which is controlled by finger pressure, and the toroid angle, $\theta_{toroid}$, which is controlled by membrane transport. The finger anchor strings introduce a fixed 25° inward offset between the finger and the local toroid tangent. Figure 5B shows the workspace generated by varying these two degrees of freedom.

In enclosure grasps, the fingertips must extend beyond the widest dimension of the object, thereby preventing distal object extraction without fingertip deflection (Fig. 5D(i))(*16*). The geometric condition for enclosure is

$$\theta_{toroid} - \theta_{bending} - \theta_{offset} < 90^{\circ}$$

where $\theta_{offset}$ = 25° is imposed by the tethered finger anchor. Figure 5D(ii) maps the workspace regions in which this condition is satisfied. Increasing $\theta_{toroid}$ requires greater finger bending—and therefore higher finger pressure—to maintain enclosure. The feasible region is additionally bounded by collision between the opposing fingertips.

Fig. 5C reports the maximum graspable cylinder diameter as a function of toroid angle and finger pressure. The feasible region is bounded by fingertip collision at small apertures and loss of enclosure when the fingertips no longer extend beyond the object boundary. Within these limits, maximum graspable diameter generally increases with toroid angle and pressure, as illustrated by limiting configurations from 10 to 50 kPa. Fingertip grasps instead rely on friction at localized contacts between the fingertips and the sides of the object. They can accommodate objects that cannot be fully enclosed, including objects with large aspect ratios, but their stability depends on the available friction. As illustrated in Fig. 5E(i), successful grasping requires the tangential load to remain within the friction limit: $-\mu F_n \leq F_t \leq \mu F_n$ where $F_n$ and $F_t$ are the normal and tangential components of the fingertip force and $\mu$ is the coefficient of friction. Figure 5E(ii) shows how membrane transport and finger bending alter fingertip force orientation and its normal and tangential components, enabling configurations with larger normal forces for more stable fingertip grasps. The body pressure regulates grasp stiffness. Once inflated, the fingers are relatively rigid compared with the compliant finger–body interface and therefore pivot primarily at their bases under external loading.

Figure 5F shows that fingertip stiffness, measured normal to the contact surface, increases from approximately 0.01 N mm$^{-1}$ at 5 kPa body pressure to approximately 0.07 N mm$^{-1}$ at 40 kPa. The system therefore provides a separation of functions: finger pressure and toroid position jointly determine grasp geometry and force alignment, whereas body pressure regulates contact stiffness. At each tested body pressure, enclosure grasps produced higher median peak pull-out forces than fingertip grasps, indicating greater grasp retention. For both grasp modes, pull-out force increased monotonically with body pressure (Fig. 5G), informing grasp-mode selection according to object geometry and task requirements(*17*, *18*).

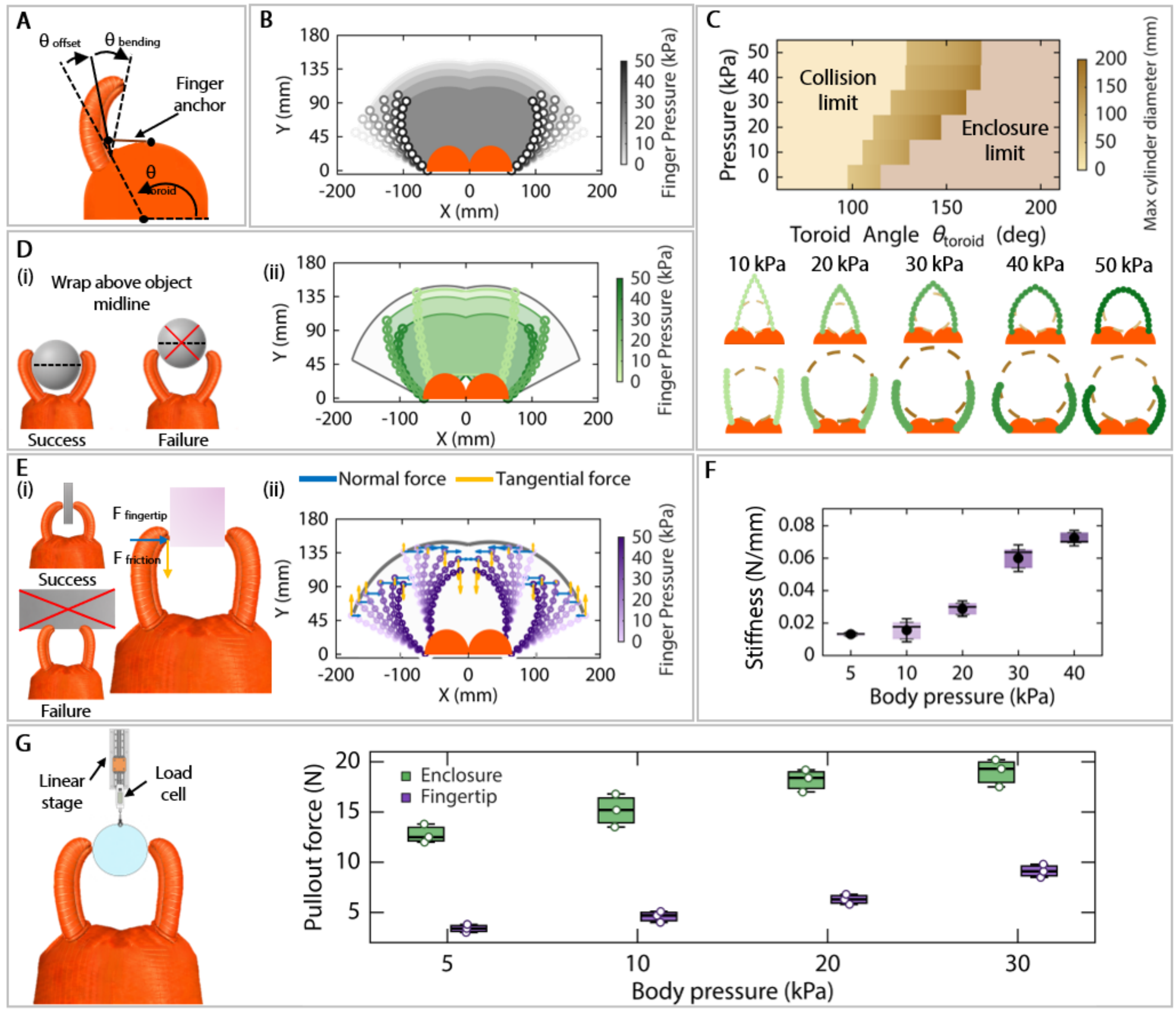

**Fig. 5. Coupled body–finger mechanics define enclosure and fingertip grasping regimes. (A)** Finger configuration is determined by bending angle $\theta_{bending}$, controlled through finger pressure, and toroid angle $\theta_{toroid}$, set through membrane cycling; passive anchor strings limit outward rotation at the finger base and define an inward offset angle $\theta_{offset}$. **(B)** Reachable fingertip workspace obtained by varying $\theta_{bending}$ and $\theta_{toroid}$. Gray trajectories and markers indicate accessible configurations, with darker shades corresponding to higher finger pressures. **(C)** Maximum diameter of a cylindrical object that can be enclosed as a function of finger pressure and $\theta_{toroid}$. The map is bounded by fingertip collision and loss of enclosure; representative configurations at 10 to 50 kPa illustrate the collision limit in the top row and enclosure limit in the bottom. **(D)** Enclosure grasping: (i) successful and unsuccessful configurations, classified by whether the fingers wrap beyond the object midline; (ii) enclosure-compatible workspace (green) within the reachable workspace boundary (gray). **(E)** Fingertip grasping: (i) successful and unsuccessful configurations and associated fingertip and friction forces; (ii) normalized normal and tangential fingertip-force components across pressure-dependent

configurations. **(F)** Effective fingertip stiffness versus body pressure. Points indicate mean values, and error bars denote ±1 SD across five trials . **(G)** Peak pull-out force versus body pressure for enclosure and fingertip grasps of a 100-mm-diameter cylinder (50 kPa finger pressure n=3 per condition). Boxes show medians and interquartile ranges while whiskers show the non-outlier range.

## Coordinated manipulation and task-specific deployment

To evaluate the robot ability to perform coordinated control of body shape, membrane position, and finger pressure we investigated its application a range of manipulation and deployment tasks (Fig. 6). In the stacking task, the robot extended toward a cylindrical object while steering the distal body to pre-shape the grasp, inflated the fingers to acquire and lift the object, and placed it on a second cylinder (Fig. 6A and movie S3). The sequence combined precise body extension and steering, wall retraction, finger inflation, and release. To assess the range of objects compatible with the two grasp modes, we conducted a qualitative retention screen using nine objects spanning masses of 21–550 g and characteristic dimensions of 16–200 mm. The set included slender, spherical, cuboidal, cylindrical, and container-like geometries; three objects were assigned to fingertip grasps and six to enclosure grasps based on their geometry. All nine objects in this graspable set were retained. Figure 6, B and C and movie S3 show representative examples. Fingertip grasps with the 120-mm fingers retained a strawberry and a plastic screw, whereas enclosure grasps retained a ball and rectangular carton with the 120-mm fingers and a plastic screw and strawberry with the 70-mm fingers. Three additional objects were selected to probe boundary conditions. The screwdriver underwent rotational slip because its offset mass distribution generated a moment that could not be resisted by the limited fingertip contacts. The beach ball exceeded the enclosure limit, preventing the fingers from extending beyond its midline. For the small cylindrical object, the fingers interdigitated rather than closing around it. These cases show that fingertip retention is sensitive to contact geometry and mass distribution, whereas enclosure grasping is limited by object size and finger closure. Each object grasp condition was attempted with the robot held horizontally. After acquisition, the body was retracted and the condition was classified as successful if the object remained captured throughout retraction or as unsuccessful if the object escaped. These tests therefore identify operating regimes and failure modes rather than estimating grasp success rates. In another demonstration, body-integrated fingers provided temporary support during growth (Fig. 6D and movie S4). Without support, the horizontally-extending body developed buckling-induced wrinkles due to the gravity moment at the base (*19*) and sagged at approximately 70 cm. After retraction, a short finger was aligned with a vertical tube using membrane cycling, inflated to grasp the tube, reducing the unsupported span and allowing the body to extend to its length of 110 cm without buckling. The robot can act as a pan–tilt platform for cameras mounted to the membrane wall. In a pipe inspection scenario, the robot extended into a pipe, then a miniature camera was transported along the membrane to the distal tip (Fig. 6E and movie S5). Membrane cycling moved the camera around the torus to pan its view while the tip location remained approximately fixed. Body steering in the vertical and horizontal directions tilted the camera view. During subsequent body motion into the pipe, coordinated membrane transport kept the camera at the tip, allowing locations within the pipe to be inspected. For controlled payload delivery, two objects were stored at different membrane-material positions within the robot tail (Fig. 6F and movie S6). The known object positions and commanded wall and tail lengths were used to determine the membrane motions required to bring each payload to the distal opening. Here, delivery was considered successful when the object reached the target region without jamming or premature release. The robot combined wall retraction, body extension, and steering to reach two separated delivery targets and release the objects sequentially. Because membrane motion and distal-body pose are decoupled, payloads stored within the same tail can be delivered at locations in the workspace using the wall and tail actuators. An additional outdoor deployment in the Mini-Mars Yard at NASA's Jet Propulsion Laboratory demonstrated target-to-target navigation and sample-container delivery in a Martian analog environment (movie S7)

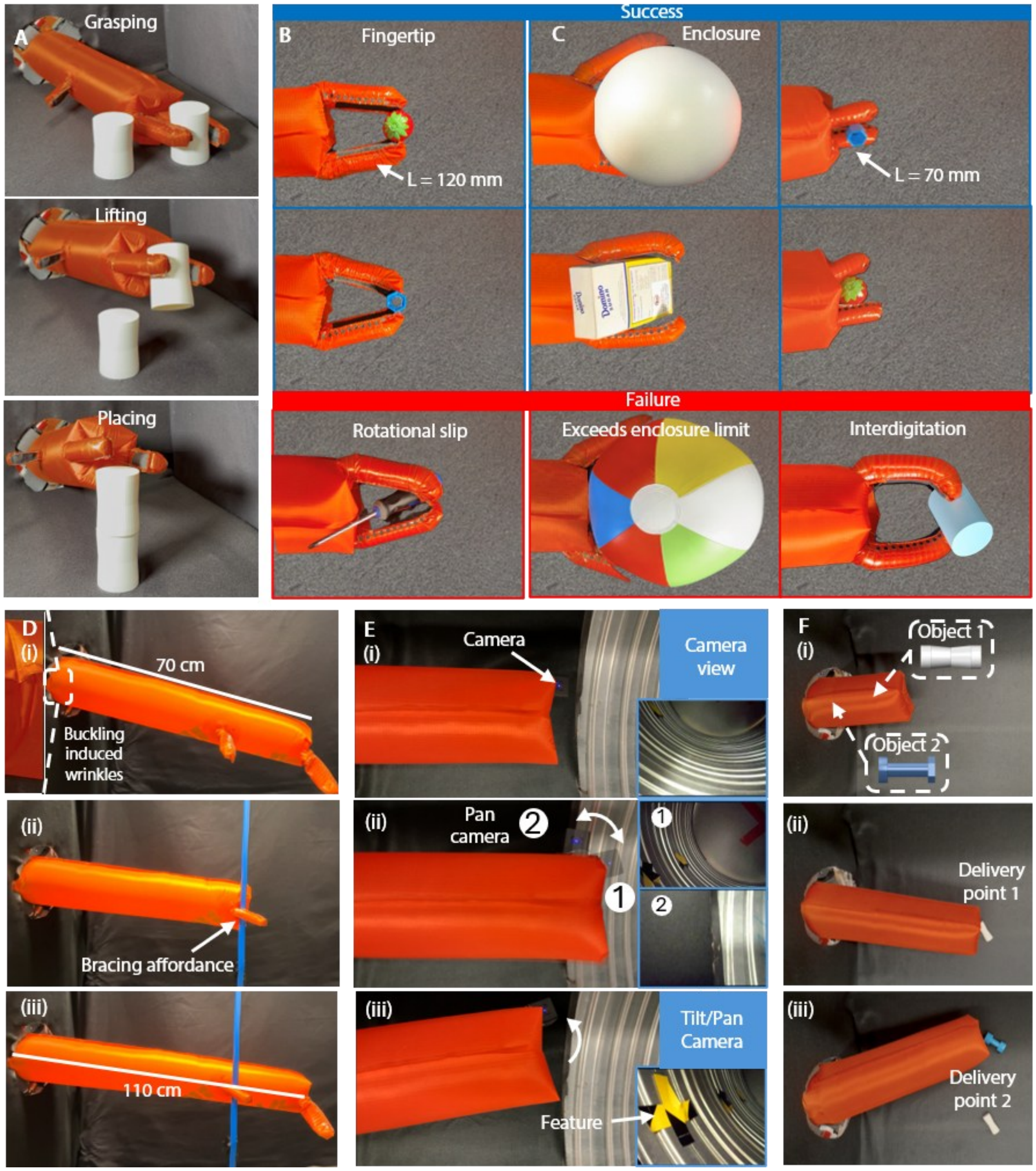

**Fig. 6. Coordinated body, membrane, and finger actuation enables manipulation and task-specific deployment. (A)** Stacking sequence showing grasp pre-shaping, acquisition, lifting, and placement on a second cylinder. **(B)** Fingertip grasps with 120-mm fingers retain a strawberry and plastic screw; a screwdriver fails by rotational slip. **(C)** Enclosure grasps retain a ball and rectangular carton with 120-mm fingers and a plastic screw and strawberry with 70-mm fingers. Failure occurs when a beach ball exceeds the enclosure limit or the fingers interdigitate around a small cylindrical object. Blue and red borders denote successful and unsuccessful retention during horizontal body retraction; each object–grasp condition was tested once. **(D)** Environmental bracing: the unsupported body develops buckling-induced wrinkles at 70 cm, whereas contact with a vertical tube enables extension to 110 cm. **(E)** A membrane-mounted camera is deployed to the distal tip and panned or tilted to inspect pipe features while remaining near the tip during body motion. **(F)** Sequential delivery of two objects to separate target locations.

## DISCUSSION

The foregoing results demonstrate that wall actuation addresses a central end-effector challenge in SERs. By metering the outer wall independently of the tail, we convert a single-degree-of-freedom growth process into a programmable material flow in which body length and membrane-material position become separately addressable states. This allows membrane-mounted tools to be deployed at the distal tip and subsequently repositioned or stowed without rigid mounts or external deployment mechanisms.

The motion-control framework provides a compact set of primitives for exploiting membrane mobility. Sequencing eversion, wall motion, membrane cycling, body extension, and steering allows the robot to regulate both distal-body configuration and the location of membrane-mounted devices. The kinematic model provides the mapping from desired tip and membrane states to actuator motion, with open-loop experiments yielding positioning errors of 5.0%–6.8% of commanded travel. Although these experiments were performed under quasi-static conditions, they establish that body motion and membrane transport can be coordinated with good repeatability and with prospects for high accuracy under closed-loop control.

While a wide variety of tools can be mounted on the robot membrane, the pleated inflatable fingers proposed here provide a number of important advantages. Lightweight textile construction allows them to remain compliant and compact during transport while producing pressure-dependent bending motion and force application after deployment. Manipulation can be accomplished through the coupling between finger and body mechanics: finger pressure regulates finger curvature, membrane position determines the grasp aperture and contact configuration, and body pressure modulates the stiffness of the finger–body interface. This separation provides a low-dimensional strategy for adapting the grasp to objects with different geometries and sizes. The same architecture also supported a variety of capabilities, including environmental bracing, camera pan-tilt pointing, sequential multi-tool deployment, and payload delivery, illustrating that wall retraction is not limited to grasping but can enable multiple forms of distal interaction.

Wall actuation entails limitations compared to conventional SERs. Wall retraction necessitates increases in hardware cost and control complexity. The number of actuators increases from one in conventional SERs to three for planar wall retraction implementations and at least four for three-dimensional implementations. In addition, wall retraction requires the external membrane to slide relative to the environment. Distributed frictional forces will increase with extended environmental contact. Wall retraction may become ineffective if the robot deploys across long paths or through constrained apertures, although conventional SER tip eversion will be preserved.

Membrane-mounted tools can nevertheless remain useful under these conditions if pre-positioned before the robot enters the constrained region. For example, a camera could be cycled through a tight opening to inspect the environment and estimate the target location. After the camera is recovered, the wall rollers and tail spool could position finger modules at the membrane location predicted to emerge at the target during growth. The robot could then traverse the opening using conventional eversion, with the tool positioned for interaction after deployment. This strategy preserves the benefits of deployable tools without requiring wall retraction during high-friction environmental contact. Future implementations could reduce frictional limitations through alternative membrane materials, improved traction management, localized reinforcement, time-varying pressure regulation, or spatially localized wall actuation.

Our prototype system was sufficient to validate the proposed concepts, but many design issues remain. Three wall rollers (plus the tail roller) are sufficient to generate the five motion control modes, but additional wall rollers may improve traction by distributing contact forces over a larger area of the membrane. (See Supplementary Information sections S2 and S3 below.) This could reduce slippage and stress concentrations and permit finer regulation of membrane motion, at the expense of additional hardware and control complexity.

In addition, the kinematic framework assumes that the membrane remains tensioned and the robot body does not buckle. It therefore does not capture configurations involving large body deformation, the effects of environmental support, or instabilities such as buckling. Extending the model to include nonlinear body mechanics, contact constraints, and body-state sensing would broaden its utility beyond the quasi-static operating regime. The grasping and functional demonstrations were also conducted over a limited set of objects and conditions. Further studies are required to quantify payload capacity, grasp success, durability, energy consumption, and performance over longer and more tortuous deployments.

The task-level controller operated open loop, without body-shape or contact feedback. The pneumatic fingers produce repeatable pressure-dependent motion, but their grasp configuration and contact force were not adjusted in response to object geometry, slippage, or deformation. Integrating shape, pressure, or contact sensing would enable feedback control of tool position and grasp force, particularly for fragile or irregular objects. Closed-loop body sensing could also compensate for membrane drift, roller slip, and construction tolerances that contribute to positioning errors in the open-loop experiments.

Finally, membrane transport need not be limited to a single tool near the distal tip. Multiple devices could be distributed along the membrane and selectively brought into the workspace, enabling sequential sensing, anchoring, manipulation, or task-specific behaviors. Realizing such systems will require scalable routing of pneumatic and electrical connections, reliable tool passage through the base mechanism, and planning methods that account for robot shape, membrane position, tool state, and environmental contact. More broadly, treating the membrane as a transport surface for functional elements may enable soft growing robots to dynamically configure not only where they reach, but also how they interact with their surroundings.

## CONCLUSIONS

This paper introduced a wall-retraction architecture that expands the role of soft everting robots from primarily locomotion into an actively controlled vehicle for manipulation. By combining independent wall and tail actuation with lightweight deployable fingers, the system can position tools near the distal workspace, adapt grasp geometry, and interact with objects while retaining compliance and deployability. Experimental results demonstrated accurate open-loop positioning together with grasping, bracing, camera deployment, and targeted payload release across a range of representative tasks.

More broadly, this work shows that useful manipulation in growing robots need not require stiff or heavy end effectors. Instead, functional elements can be carried by the membrane itself and brought into the workspace. Future systems that incorporate feedback sensing, improved traction, and multiple distributed tools could build on this approach to support more autonomous and versatile interaction in cluttered, confined, and difficult-to-access environments.

### MATERIALS AND METHODS

## Robot architecture and actuation

The robot consisted of a soft everting body, a tail spool actuator, and a wall retraction device composed of four independently driven roller actuators (Fig. 1). The everting body was fabricated from TPU-coated nylon fabric with a thickness of 0.1 mm and an areal density of 140 g $m^{-2}$ (AdventureXpert, USA). The fabric was formed into a pressurized cylindrical membrane with a nominal diameter of 127 mm and a deployable length of 1100 mm.

The base actuation system was designed to preserve an open central lumen, allowing internal routing of deployable tools, pneumatic lines, and transported objects. The tail was stored on an independently actuated spool with an effective diameter of 36.6 mm, which regulated the amount of material supplied to the inner body during eversion. Four independently driven captive-roller sets (44.32 mm effective diameter), arranged around the membrane circumference at the robot base, metered outer wall motion. The tail spool and wall retraction rollers were driven by 30W Maxon Motors (model #539473) with encoder feedback of 4096 counts per revolution with a 190:1 gear ratio. Two passive rollers, with a 39.28 mm diameter, maintained contact between the membrane and active rollers and guided the membrane through the device without obstructing the lumen.

Body pressure was regulated using an electronic pressure regulator (QB3, Proportion-Air Inc) over a range of 0-40 kPa and was used to drive eversion and tune the body mechanical impedance. Pressure for each finger set (small and large) was supplied through separate pneumatic lines using the same type of electronic pressure regulators over a range of 0-50 kPa.

Tail and wall encoder counts were converted into linear membrane displacements using calibrated roller and spool diameters. Average wall commands were generated by actuating all wall sectors together, whereas differential wall commands were generated by imposing different length changes on selected wall sectors. Membrane cycling and body extension commands were implemented by coordinating tail spool motion with the average wall displacement. Detailed actuator specifications, roller geometry, membrane routing, and displacement calibration procedures are provided in the Supplementary Materials.

## Kinematic control

The kinematic controller was implemented as an open-loop command generator that converted desired distal displacements into tail-spool and wall-roller commands. The four wall-sector displacements and the tail displacement were first computed in body-length space and then converted into motor rotations. The membrane-material correction described in Eq. (1) was applied to compensate for predicted membrane-point drift relative to the distal tip. Motor commands were executed using synchronized position control at 100 Hz.

The controller was evaluated using the trajectories in Fig. 3B(i–iv). Planar positioning was tested using waypoint sequences in the transverse plane, including the “H” and “+” shaped trajectories shown in Fig. 3B(i) and Fig. 3B(ii), respectively. Axial body extension and retraction were tested along the robot’s longitudinal direction, as shown in Fig. 3B(iii), whereas membrane cycling was tested by transporting the tracked membrane point while maintaining a nominally constant distal pose, as shown in Fig. 3B(iv). For each trajectory, a sequence of waypoints was commanded, and the robot was allowed to reach a steady configuration before the next command. Each trajectory was repeated five times.

Distal motion was measured using two 6-DoF electromagnetic tracking sensors (trakSTAR, Northern Digital Inc.). The sensor used to measure $P_E$ was attached at the geometric center of the distal torus, whereas the sensor used to measure $P_M$ was attached to the tracked membrane point at $S_M$. The transmitter was positioned approximately 520 mm from the workspace, and tracking data were acquired at 80 Hz. The tracker coordinate frame was registered to the robot base frame using a rigid transformation. Position data were projected onto the relevant plotting planes: X/Y for planar positioning and Z/Y or Z/X for axial and membrane-cycling motions.

For each waypoint, the Euclidean position error, $e$, was calculated as $e = \|P_{Meas} - P_{Cmd}\|$, where $P_{Meas}$ is the measured position after settling and $P_{Cmd}$ is the commanded waypoint position. For the visualizations in Fig. 3B(i–iv), the trajectories were normalized by the maximum commanded displacement along each axis to allow comparison across motion types. The mean trajectory and one standard deviation across the five repeated trials are reported in Fig. 3B(i–iv). The full derivation of the membrane-transport model, sign conventions, actuator-space mapping, and controller pseudocode is provided in the Supplementary Materials.

## Fabrication and integration of pleated inflatable fingers

The fabric layers were laser cut from the same material used for the everting body. The top layer was folded into the pleated configuration and aligned with the smooth bottom layer. The two layers were sewn along the perimeter, leaving an internal cavity for the bladder. The distal end was closed by sewing box pleats to form the end cap. An airtight bladder was fabricated from 0.05 mm thick TPU film, heat sealed around its perimeter, inserted inside the fabric shell, and connected to a flexible pneumatic tube. The bladder was not bonded to the fabric shell, except at the tubing interface.

The fingers were attached to the everting membrane by bonding the proximal finger base to the TPU-coated side of the robot body using heat pressing. The active pleated region was left free to bend during inflation. The bonding area was kept approximately planar to allow the fingers to pass through the wall retraction device during membrane transport. Pneumatic tubes were routed along the inside of the membrane and connected to independent pressure lines. Fingers with a length of 120 mm and $N_p$ = 10, 12, 14, and 16 pleats were fabricated. A geometrically scaled 70-mm-long version of the $N_p$= 12 finger was fabricated for length-dependent comparisons. Detailed cutting patterns, bonding parameters, and tubing routing are provided in the Supplementary Materials.

## Finger characterization

Finger bending was characterized as a function of inflation pressure, finger length, and pleat number. Each finger was clamped at its proximal base and inflated from 0 to 50 kPa in 10 kPa increments, with a 10 s settling time at each pressure. Side-view images were acquired with a camera positioned orthogonally to the bending plane and calibrated using a marker of known length. Centerline markers were manually digitized and fitted with a cubic smoothing spline; under a constant-curvature assumption, a circular arc was fitted to the reconstructed centerline and the bending angle was calculated as $\theta = \kappa L$. Fingers with $N_p$= 10, 12, 14, and 16 pleats were tested, while the length-dependent and blocking-force comparisons used $N_p$= 12 fingers with $L$ = 70 and 120 mm. Blocking force was measured using a SAUTER FK10 digital force gauge (10 N capacity), with the distal tip placed against the probe without preload and the probe oriented approximately normal to the expected motion. The force was recorded after 10 s of stabilization.

## Body–finger coupling and grasping experiments

The body–finger grasping interface was characterized by combining measured finger shapes with the distal torus geometry. Finger shapes obtained at different pressures were registered to the torus coordinate frame. The swept area of the fingers was then used to estimate the regions compatible with object capture and finger-collision limits. Workspace maps were generated over finger pressures of 0–50 kPa and torus angles of 0–180° measured from horizontal.

Two grasping regimes were evaluated. Enclosure grasps were classified as configurations in which the object could be confined within the toroidal body and the inflated fingers, with the fingers wrapping around or beyond the object's widest region. Fingertip grasps were classified as configurations in which contact occurred mainly near the distal portions of the fingers without full enclosure. For cylindrical-object analysis, the maximum admissible object diameter was estimated from the registered finger shapes and torus aperture before finger collision at an empirically acquired angle limit of 100°. After grasp acquisition, a cylindrical object was connected to a digital force gauge through a monofilament line aligned with the robot's longitudinal axis. The force gauge was mounted on a motorized linear stage that displaced the object at 2 mm $s^{-1}$ until release occurred. The maximum force recorded immediately before release was defined as the pull-out force. Tests were conducted at body pressures of 5, 10, 20, 30 and 40 kPa while maintaining the finger pressure at 50 kPa. Each grasp configuration and body-pressure condition was tested in three trials.

Distal body stiffness was measured by applying indentation displacements to the toroidal region at body pressures of 5–40 kPa with a 6.35 mm diameter probe coated with high friction tape (3M, TB641). Reaction force was measured using an ATI Nano17 six-axis force/torque sensor mounted on a motorized linear stage and stiffness was computed as the slope of the force–displacement curve over the linear range 0–20 mm. All force data were baseline-corrected.

## Coordinated finger and membrane tasks

The same membrane-flow architecture was evaluated in object grasping, anchoring, deployable sensing, and object delivery demonstrations. For object grasping, the robot first grew toward the target, transported the selected fingers to the distal tip by membrane cycling, adjusted the grasp aperture through wall and torus motion, and inflated the fingers to acquire the object. Objects were selected to span different sizes, masses, and geometries, including cylinders, cubes, prisms, spheres and irregular objects. A trial was considered successful when the robot acquired the object, maintained the grasp during the prescribed manipulation, and released it on command. For anchoring, the shorter 70 mm long inflatable finger was transported to the distal region, brought into contact with an external support using wall retraction and steering, and inflated to 50 kPa to create a temporary anchor during further extension. Buckling was defined as a visible transverse collapse. For deployable sensing, a miniature wireless camera (iWifiCam Wireless Camera, Amazon) with integrated LED illumination was transported along the membrane, deployed near the distal region and repositioned using coordinated membrane motion. For object delivery, objects were transported inside the hollow body and released at locations by coordinating eversion, inversion, cycling, and steering.

## Statistical analysis

Experimental data were processed using MATLAB. Encoder measurements were converted into tail and wall displacements using calibrated roller and spool diameters. Motion-tracking data were used to compute distal tip trajectories, steering angles, and waypoint errors. Image data were used to extract finger centerlines, bending angles, and workspace maps. Force data from blocking-force, pullout, and stiffness experiments were baseline-corrected. Unless otherwise stated, experiments were repeated $n = 5$ times, and data are reported as mean ± 1 standard deviation. For each pressure or displacement condition, measurements were taken only after the system reached a steady state. Outliers or failed trials were excluded only when caused by identifiable experimental errors, such as leakage or sensor data loss.

**References**

1. E. Del Dottore, A. Mondini, N. Rowe, B. Mazzolai, A growing soft robot with climbing plant–inspired adaptive behaviors for navigation in unstructured environments. Sci. Robot. 9, eadi5908 (2024).

2. J. D. Greer, L. H. Blumenschein, R. Alterovitz, E. W. Hawkes, A. M. Okamura, Robust navigation of a soft growing robot by exploiting contact with the environment. Int. J. Robot. Res. 39, 1724–1738 (2020).

3. M. Wooten, C. Frazelle, I. D. Walker, A. Kapadia, J. H. Lee, "Exploration and inspection with vine-inspired continuum robots" in 2018 IEEE International Conference on Robotics and Automation (ICRA) (IEEE, 2018), pp. 5526–5533.

4. M. M. Coad, L. H. Blumenschein, S. Cutler, J. A. R. Zepeda, N. D. Naclerio, H. El-Hussieny, U. Mehmood, J.-H. Ryu, E. W. Hawkes, A. M. Okamura, Vine robots. IEEE Robot. Autom. Mag. 27, 120–132 (2019).

5. C. Laschi, B. Mazzolai, M. Cianchetti, Soft robotics: Technologies and systems pushing the boundaries of robot abilities. Sci. Robot. 1, eaah3690 (2016).

6. L. H. Blumenschein, M. M. Coad, D. A. Haggerty, A. M. Okamura, E. W. Hawkes, Design, modeling, control, and application of everting vine robots. Front. Robot. AI 7, 548266 (2020).

7. N. D. Naclerio, A. Karsai, M. Murray-Cooper, Y. Ozkan-Aydin, E. Aydin, D. I. Goldman, E. W. Hawkes, Controlling subterranean forces enables a fast, steerable, burrowing soft robot. Sci. Robot. 6, eabe2922 (2021).

8. P. A. der Maur, B. Djambazi, Y. Haberthür, P. Hörmann, A. Kübler, M. Lustenberger, S. Sigrist, O. Vigen, J. Förster, F. Achermann, et al., "Roboa: Construction and evaluation of a steerable vine robot for search and rescue applications" in 2021 IEEE 4th International Conference on Soft Robotics (RoboSoft) (IEEE, 2021), pp. 15–20..

9. H. Tsukagoshi, N. Arai, I. Kiryu, A. Kitagawa, Tip growing actuator with the hose-like structure aiming for inspection on narrow terrain. Int. J. Autom. Technol. 5, 516–522 (2011).

10. S. K. Talas, B. A. Baydere, T. Altinsoy, C. Tutcu, E. Samur, Design and development of a growing pneumatic soft robot. Soft Robot. 7, 521–533 (2020).

11. H. Yong, F. Xu, C. Li, H. Ding, Z. Wu, “Design and modeling of a nested bi-cavity-based soft growing robot for grasping in constrained environments” in 2024 IEEE International Conference on Robotics and Automation (ICRA) (IEEE, 2024), pp. 2346–2352.

12. E. M. DeVries, J. Ferlazzo, M. Ugur, L. H. Blumenschein, Transport and Delivery of Objects with a Soft Everting Robot. IEEE Robot. Autom. Lett. 11, 2935–2942 (2026).

13. N. G. Kim, N. J. Greenidge, J. Davy, S. Park, J. H. Chandler, J.-H. Ryu, P. Valdastri, External Steering of Vine Robots via Magnetic Actuation. Soft Robot. 12, 159–170 (2025).

14. D. Stewart, A Platform with Six Degrees of Freedom. Proc. Inst. Mech. Eng. 180, 371–386 (1965).

15. K. M. Lynch, F. C. Park, Modern Robotics: Mechanics, Planning, and Control (Cambridge University Press, Cambridge, 2017).

16. W. S. Howard, V. Kumar, On the Stability of Grasped Objects. IEEE Trans. Robot. Autom. 12, 904–917 (1996).

17. D. Prattichizzo, J. C. Trinkle, “Grasping” in Springer Handbook of Robotics, B. Siciliano, O. Khatib, Eds. (Springer, Cham, ed. 2, 2016), pp. 955–988.

18. R. Newbury, M. Gu, L. Chumbley, A. Mousavian, C. Eppner, J. Leitner, J. Bohg, A. Morales, T. Asfour, D. Kragic, D. Fox, A. Cosgun, Deep Learning Approaches to Grasp Synthesis: A Review. IEEE Trans. Robot. 39, 3994–4015 (2023).

19. C. McFarland, M. M. Coad, “Collapse of straight soft growing inflated beam robots under their own weight” in 2023 IEEE International Conference on Soft Robotics (RoboSoft) (IEEE, 2023), pp. 1–8.

20. R. C. Hibbeler, Engineering Mechanics: Statics (Pearson, Hoboken, ed. 14, 2016), pp. 439–440.

21. P. H. Nguyen, W. Zhang, Design and computational modeling of fabric soft pneumatic actuators for wearable assistive devices. Sci. Rep. 10, 9638 (2020).

**Acknowledgments:** We gratefully acknowledge the support of Fabely Moreno Ferrer on prototype construction, experimental procedures and demonstrations. We also extend our appreciation to Alessandra De Maio, Gwen Reimer, and Emilia Mann for assistance in the soft finger construction and implementation process, and Irene Mannari for supporting the preparation of Figure 1. We extend our gratitude to the Harvard University REEF Makerspace for access to prototyping equipment used throughout the project as well as NASA’s Jet Propulsion Laboratory for use of their testing facilities.

**Funding:** N.B.P was supported by the National Aeronautics and Space Administration (NASA) Space Technologies Graduate Research Opportunities Fellowship grant #80NSSC23K1226. N.P. was supported by Erasmus+ mobility grant from the Scuola Superiore Sant'Anna.

**Author contributions:** Conceptualization: N.B.P., N.P., and R.D.H. Methodology: N.B.P., N.P., and R.D.H. Investigation: N.B.P. and N.P. Software: N.B.P. Formal analysis: N.B.P. and N.P. Visualization: N.B.P. and N.P. Writing—original draft: N.B.P. and N.P. Writing—review and editing: N.B.P., N.P., M.C., and R.D.H. Supervision: M.C. and R.D.H. Funding acquisition: M.C and R.D.H.

**Competing interests:** R.D.H. has a financial interest in RightHand Robotics inc. and is a named coinventor on the following US Patents on robot hands and tactile sensing: 11,338,436; 11,173,602; 10,488,284; 9,625,333; and 8,231,158.

**Data and materials availability:** All data needed to evaluate the conclusions in the paper are present in the paper and/or the Supplementary Materials.

**Supplementary materials**

**Movie S1: wall-retraction device enabled motions.** Demonstration of coordinated tail and wall actuation for eversion, wall retraction, membrane cycling, body extension, and steering.

**Movie S2: coordinated body and tool actuation and sequential multi-tool deployment.** Sequential deployment, actuation, and retraction of membrane-mounted cameras and inflatable finger sets.

**Movie S3: object stacking and manipulation.** Coordinated grasp pre-shaping, acquisition, lifting, manipulation, and stacking of objects using body and finger actuation.

**Movie S4: environmental bracing enabled by body mounted fingers.** Deployment of an inflatable finger as a temporary environmental support to increase the robot's unsupported extension.

**Movie S5: body actuated camera panning and tilting for inspection.** Deployment and repositioning of a membrane-mounted camera for distal inspection using membrane cycling and body steering.

**Movie S6: controlled multi-object disgorgement with wall retraction.** Sequential transport and release of multiple payloads at distinct target locations using coordinated wall and tail actuation.

**Movie S7: deployment in a Martian analog environment.** Outdoor deployment in the Mini-Mars Yard at NASA's Jet Propulsion Laboratory, demonstrating multi-motion mode navigation between target locations and delivery of a sample-container tube to a designated drop-off site.

## S1. System architecture and hardware implementation

The prototype robot has two subsystems: a compliant membrane body and a rigid base station (Fig. S1). The membrane is a cylindrical everting tube that extends beyond the base to interact with the environment. Appendages such as the pleated fingers described in Sections S7 and S8 attach to the membrane wall. Material not currently deployed is stored in a reservoir located between the base and the wall retraction device, where axial buckling compacts long segments of membrane into a small volume through pleating.

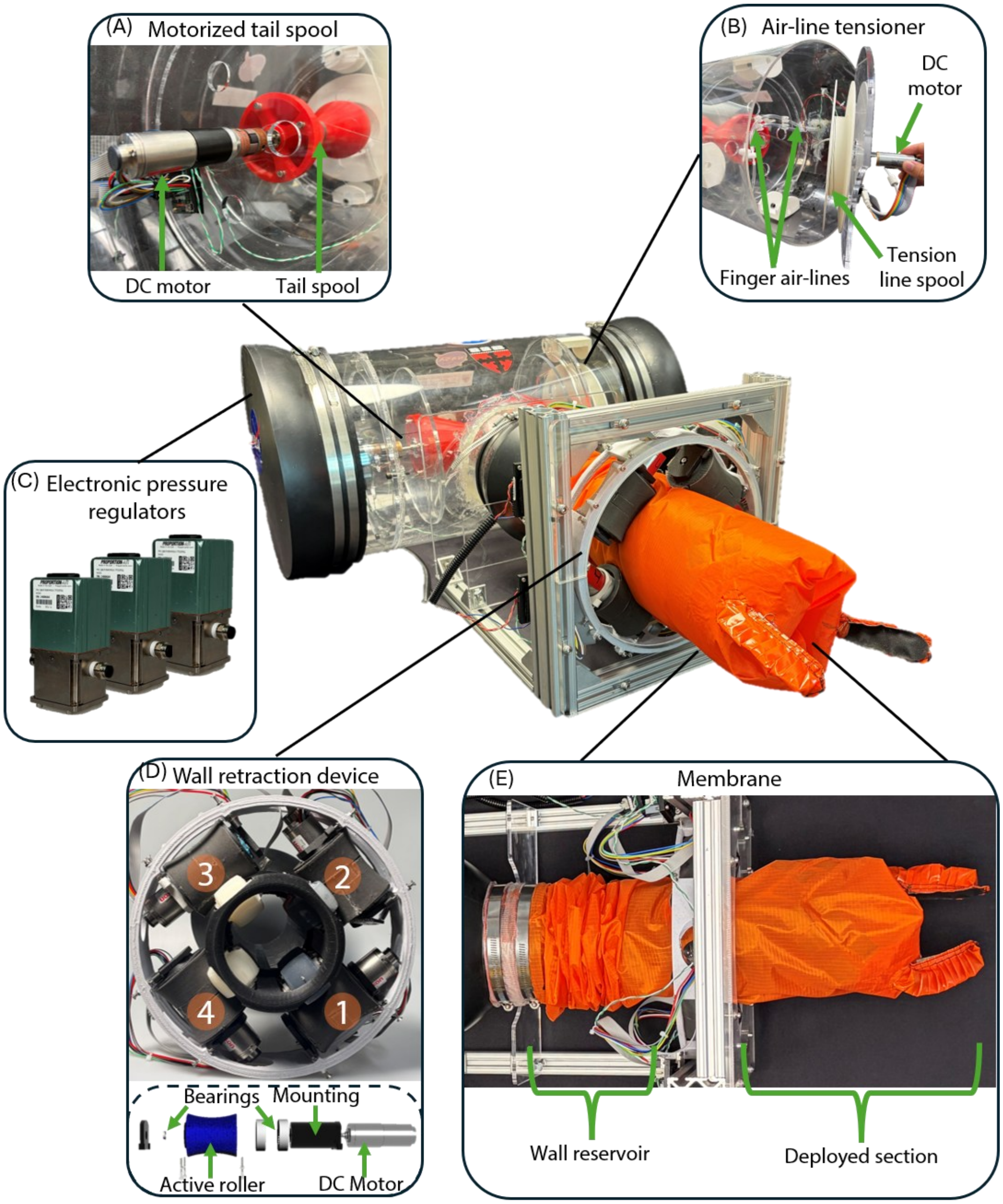

**Fig. S1. Hardware implementation. (A)** Motorized tail spool. (B) Air-line tensioner. (C) Electronic pressure regulators. (D) Wall retraction device with four active roller assemblies (numbered 1–4), shown with an exploded view of a single roller. (E) Membrane in its deployed and reservoir states, with the pleated reservoir visible between the base and the wall retraction device.

The base station houses the electromechanical and pneumatic hardware. A hyperboloid-shaped tail spool, driven by a geared DC motor, actuates the membrane tail. A second spool regulates tension in the air-lines that route pressure from the regulators to body-mounted pneumatic elements, preventing damage during motions that require wall retraction. Electronic pressure regulators hold body pressure at a programmed set point. A wall retraction device with four active rollers drives the deployed length of the wall. Each roller is mounted concentrically around its DC motor on needle bearings.

We separate pressure and motion control. Body pressure is held above the eversion threshold (≈0.25 kPa) to keep the membrane in tension. Motion is then governed by tail and wall roller motor positions rather than by flow-rate-dependent growth, transferring control authority from pneumatics to precise motor control.

## S2. Wall retraction device design

We developed a wall retraction system capable of exerting large forces on the membrane material while also preserving airtightness and preventing unwanted slip. Here, we use a captive multi-roller configuration that uses the Capstan effect to set effective holding forces: four active rollers mounted on an outer frame interlock with eight passive rollers attached to an inner frame (Fig. S2). At each membrane motor interface, wall material passes between two passive rollers coated in silicone (Dragon Skin 30A) and an active roller covered in high friction double-sided adhesive (3M TB641).

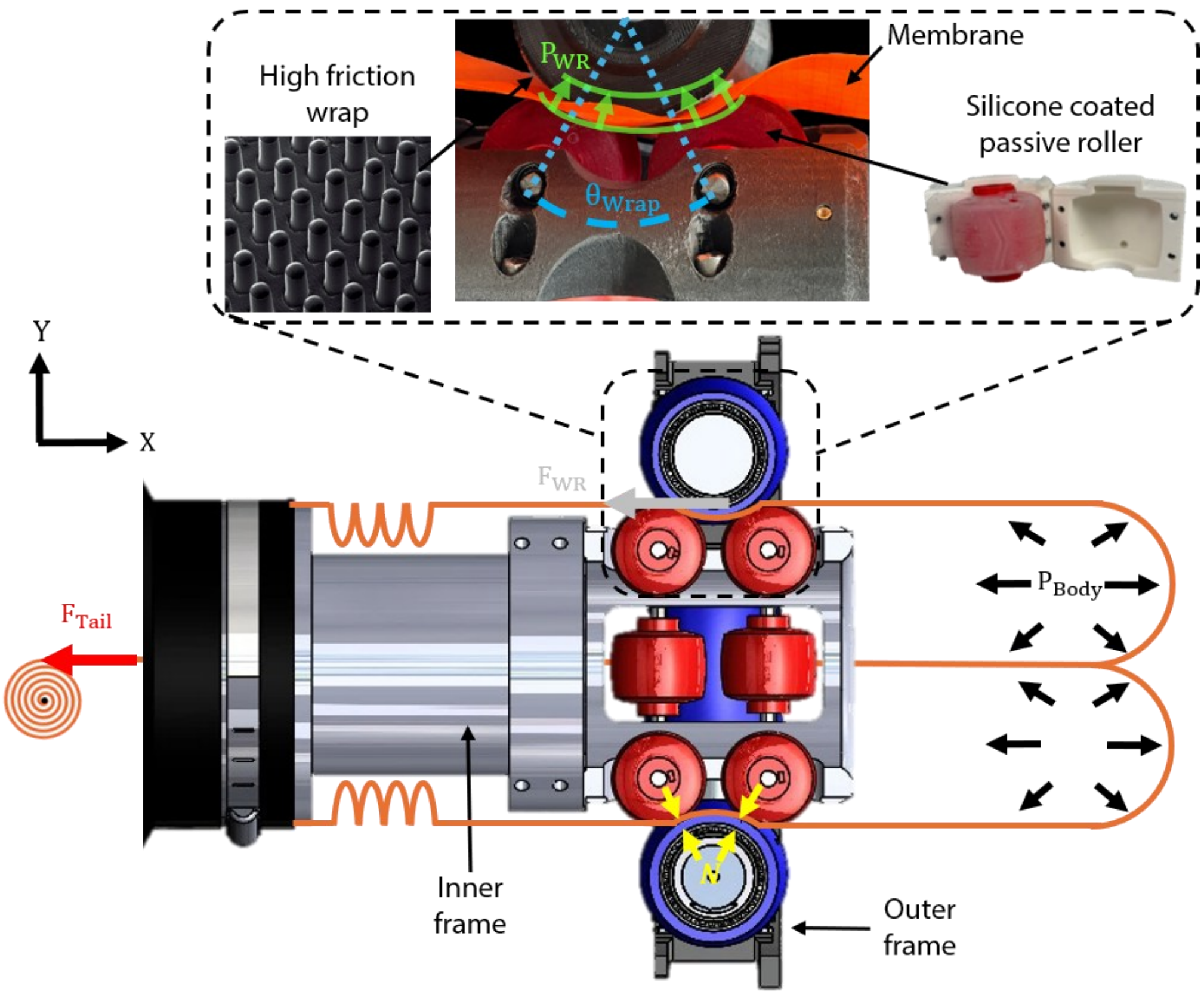

**Fig. S2. Wall-retraction module design and membrane force transmission.** Motor-driven active rollers (blue) transmit wall-retraction traction $P_{WR}$ to the membrane (orange) by pressing it against silicone-coated passive rollers (red). A high-friction wrap on the active rollers increases traction, while the compliant silicone coating on the passive rollers maintains contact forces despite fabrication and alignment imperfections. The upper inset details the roller–membrane interface and the membrane wrap angle $\theta_{wrap}$. The main schematic shows the inner and outer frames and the forces acting on the membrane: internal body pressure $P_{Body}$, tail-spool tension $F_{Tail}$, normal roller contact force $N$, and resultant wall-retraction force $F_{WR}$. Black arrows indicate membrane loading produced by body pressure, and the orange line traces the membrane path through the roller assembly.

This membrane material routing forms a wrap angle $\theta_{Wrap}$ with respect to the active roller. From a force balance on the membrane in the $x$-direction (Fig. S2) we can determine the holding force $F_{WR}$ required to prevent slipping given a known internal body pressure $P_{Body}$ and tail tension $F_{Tail}$

$$F_{Tail} = P_{Body}A_{Body} - 4F_{WR}. \tag{S1}$$

Here, $A_{Body}$ is the internal cross-sectional area of the inflated robot body. For a circular body geometry, $A_{Body} = \pi R_{Body}^2$, where $R_{Body}$ is the inflated body radius. Thus, $P_{Body}A_{Body}$ represents the axial force exerted by the internal pressure, $P_{Body}$, on the robot torus. This relationship holds whenever the motion is quasi-static. We evaluate during body retraction, which represents the highest operating force scenario. We use a capstan relation (*20*) to model the change in $F_{WR}$ with the coefficient of friction between the active roller and the membrane $\mu$, contact angle $\theta_{Wrap}$, traction on the membrane $P_{WR}$, and contact area $A_{Contact}$

$$F_{WR} \leq P_{WR}A_{Contact}\left(e^{\mu\theta_{Wrap}} - 1\right). \tag{S2}$$

By substituting Eq. (S2) into Eq. (S1) we can solve for the captive roller design variables that allows us to prevent slipping given a known internal pressure

$$F_{Tail} + 4P_{WR}A_{Contact}\left(e^{\mu\theta_{Wrap}} - 1\right) \geq P_{Body}A_{Body}. \tag{S3}$$

## S3. Capstan model validation

To characterize how wrap angle affects the maximum retraction force the roller-membrane interface can sustain before slipping, we constructed an experimental setup in which one of two passive rollers could be repositioned along an arc around a central active roller to set a discrete wrap angle $\theta_{wrap}$. This is compared to a second configuration with a single roller (0° wrap) as a baseline (Fig. S3, A). This bench setup was mounted on a universal testing machine (Instron, model 68TM-10), with the membrane routed over the roller assembly and pulled at a constant displacement rate (0.5 mm/sec) while recording force and displacement. Each wrap angle condition (60°, 75°, 90°, 120°, 150°, 180°, 210°, 240°, plus the 0° single-roller baseline) was tested three times. The membrane was returned to a slack state between each trial. The slip force for each trial was defined as the force at the first prominent peak in the force-displacement curve, identified using a peak-detection algorithm (MATLAB *findpeaks*) with prominence and minimum-distance thresholds tuned per condition to exclude local peaks from sensor noise. The peak selection was verified through visual inspection of each trace.

Mean slip force increased monotonically with wrap angle, from 4.07 ± 1.63 N at 0° to 334.91 ± 36.78 N at 240° (Fig. S3B), consistent with capstan friction behavior. To benchmark the measurements against theory, we fit the classical capstan equation (S2) and compared it to the experimental data. The traction value $P_{WR}A_{Contact}$ (22.58 N) was acquired by performing a compression test where a single passive roller pressed against the membrane and active roller on the setup described above. An additional force gauge (Nextech DFS500) was used to find the compression force corresponding to pullout forces measured in our full system. Assuming that each roller contact provides half of the traction, we multiplied the resultant by 2 to obtain the final value. On the other hand, μ = 0.65 was estimated by attaching a free membrane end to a load cell (ATI Nano 17), placing a known mass on top of the membrane and the high friction active roller wrap and then measuring the force at slip. The Capstan model closely tracked the measured slip forces across the full tested range of wrap angles, supporting its use as a design tool for selecting wrap angles to achieve a target holding force given desired operating pressures.

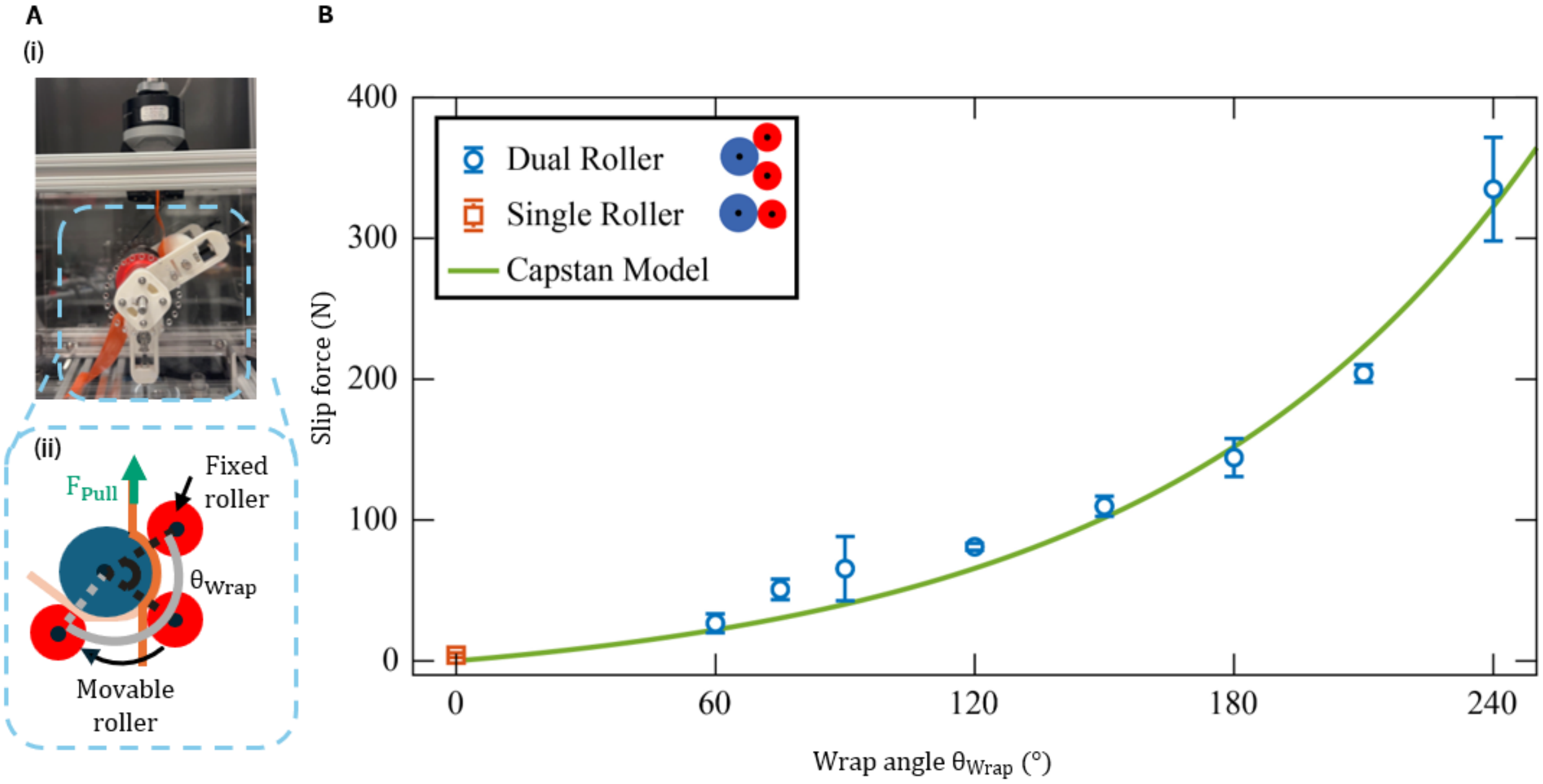

**Fig. S3. Experimental characterization of slip force as a function of wrap angle.** (A) Universal testing machine (Instron 68TM-10) implementation (top) and schematic (bottom) of the test setup. The membrane wraps around a driven roller (blue) and two passive guide rollers (red). The lower roller is repositioned to vary the membrane wrap angle, $\theta_{wrap}$, while the pulling force, $F_{pull}$, is applied vertically to one free end of the membrane. (B) Slip force as a function of $\theta_{wrap}$. Markers indicate the mean and error bars one standard deviation for the dual-roller configuration (blue circles) and the single-roller baseline at $\theta_{wrap}$ = 0° (orange square), with $n = 3$ trials per condition. The green curve represents the capstan friction model described in Eq. S2, using a holding traction, $P_{WR}A_{Contact}$ = 22.58 N, and a coefficient of friction μ = 0.65.

## S4. Membrane mechanical properties

The robot membrane's (30 Denier TPU Coated Ripstop Nylon) orthotropic mechanical response was characterized under cyclic uniaxial tension loading. Rectangular coupons (150 mm long and 75 mm wide) were cut at 0°, 45°, and 90° relative to the warp direction and tested on an Instron universal testing machine at a constant rate of 15 mm/min to a peak load of 20 N (one coupon per orientation). Each coupon was subjected to four load–unload cycles. For our analysis we extracted cycles 2–4 of the material loading branch. Then we re-zeroed to remove accumulated slack, and averaged to obtain a single representative engineering stress–strain curve per orientation. Young's moduli were fit by linear regression over the 5-25% peak-stress window, where the response was seen as approximately linear. This yielded $E_1$ = 427 MPa (warp), $E_2$ = 141 MPa (weft), and $E_{45}$ = 96 MPa ($45^o$ bias direction). While this particular material has not been characterized in the literature to the best of our knowledge, the results are consistent with similar ripstop nylons of varying denier values (*21*).

## S5. Inverse kinematic model derivation

The geometric definitions and loop-closure relations used in the inverse kinematic model are shown in Fig. S4.

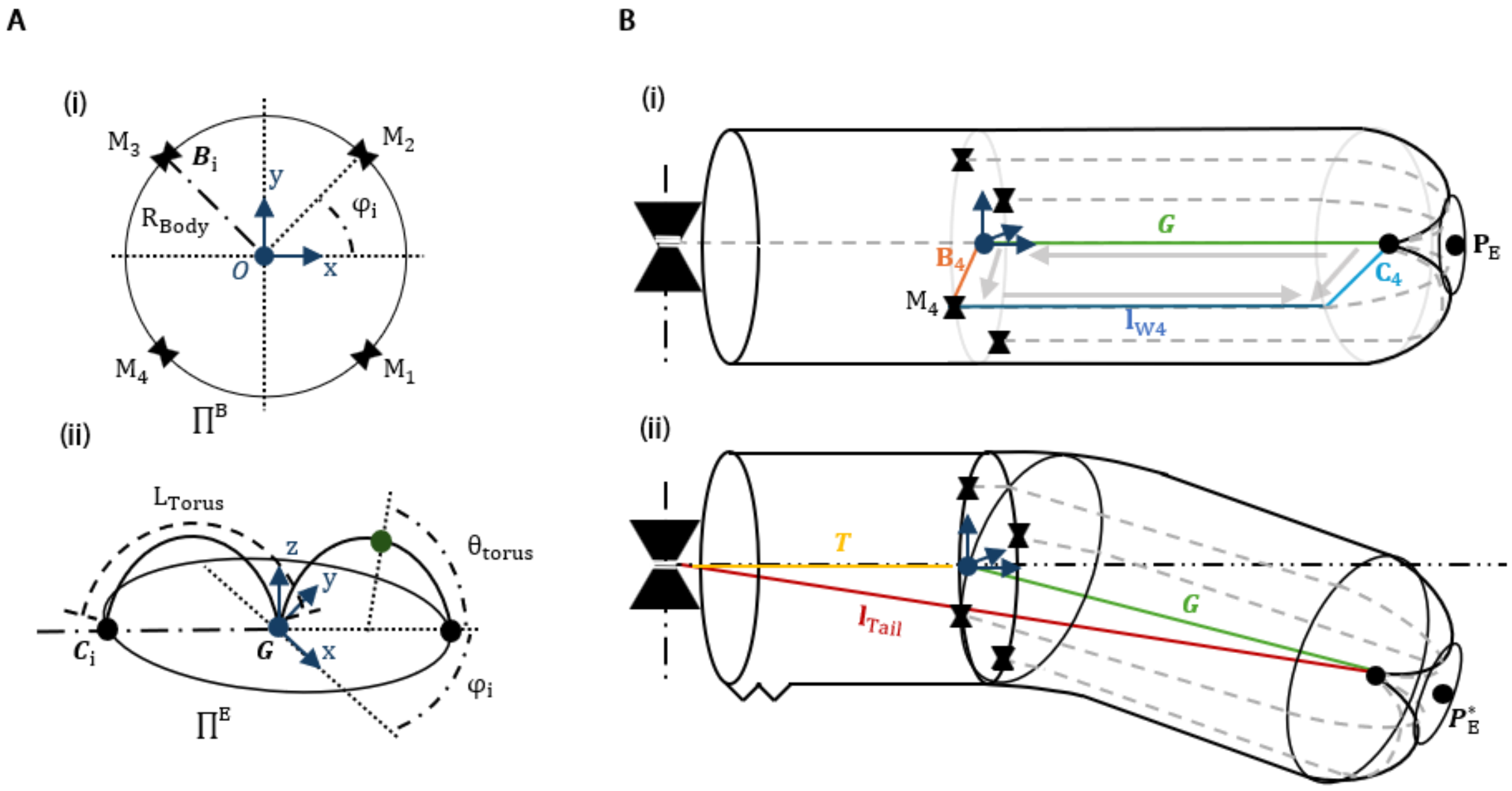

**Fig. S4. Geometric definitions and loop-closure relations for the kinematic model.** (A) Planar virtual-platform and coordinate definitions. (i) The base-platform plane, $\Pi^B$, is defined by the wall-motor locations, $M_i$. The base frame $O$ origin is at the center of the motors. (ii) The distal platform, $\Pi^E$, is defined at the intersection of the cylindrical wall and the half-torus distal end cap. The local coordinate frame is at the center of the torus. (B) Virtual-string loop closures. (i) Representative wall-string kinematic loop closure connecting the base virtual joint ($i = 4$), $B_4$, to the distal virtual joint, $C_4$, through the vector $\boldsymbol{l}_{W4}$. (ii) Tail-string closure defined by the fixed tail-motor position, **T**, the tail vector, $\boldsymbol{l}_{Tail}$, and $\boldsymbol{G}$. Dashed outlines represent the virtual strings used to model the robot body.

*Note: a nomenclature table is provided at the end of this section.*

Given the commanded distal-tip position $\boldsymbol{P}_E^*$, commanded membrane-material coordinate $S_M^*$, and current motor encoder state $\boldsymbol{q}_0$, the inverse model described in this section determines the five actuator targets $\boldsymbol{q}^*$. Through the calibrated actuator-to-length mapping and the known initialization state, $\boldsymbol{q}_0$ defines the current geometric span vector $\boldsymbol{L}_0$ and the current tagged-membrane coordinate $S_{M,0}$. This takes the form:

$$(\boldsymbol{P}_E^*, S_M^*; \boldsymbol{q}_0) \xrightarrow[Kinematics]{Inverse} \boldsymbol{q}^*$$

This mapping is broken down in the following steps as:

1. **Distal-platform configuration:** determine the constrained distal-platform orientation and center from the commanded tip position.

2. **Geometric spans:** calculate the four-wall virtual-span lengths and the tail virtual-span length required for that configuration.

3. **Membrane transport:** predict the passive membrane displacement and calculate the additional cycling displacement required to reach the commanded membrane coordinate.

4. **Actuator mapping:** combine the shape and cycling displacements and convert the resulting membrane travel into synchronized encoder targets.

The model assumes that (i) the membrane remains taut and inextensible, (ii) the body does not buckle, (iii) the rollers maintain no-slip contact with the membrane, and (iv) the roller connections can be represented as spherical virtual joints. In addition, we assume that the centerline joining the base origin $\mathbf{O}$ to the distal tip passes through the center of $\Pi^E$, and the distal plane does not rotate about this centerline. These two geometric constraints make distal orientation dependent on distal position rather than an additional command. This model applies to quasi-static motion before buckling or loss of roller traction.

**S5.1 Coordinate spaces and geometric definitions**

The base frame $\Pi^B$ has origin $\boldsymbol{O}$ at the intersection of the undeformed body axis and the base plane. Its axial unit vector is

$$\mathbf{n}_0 = [0 \quad 0 \quad 1]^T.$$

The distal frame $\Pi^E$ is centered at $\mathbf{G}$ and has normal $\mathbf{n}_E$. The rotation matrix $\mathbf{R} \in SO(3)$ maps vectors expressed in $\Pi^E$ to $\Pi^B$. The four wall rollers are located at the virtual joint locations

$$\mathbf{B}_i = \begin{bmatrix} R_{Body}\cos\varphi_i \\ R_{Body}\sin\varphi_i \\ 0 \end{bmatrix}_{\Pi_B}, \qquad \varphi_i = -\frac{\pi}{4} + (i-1)\frac{\pi}{2}, \qquad i = 1, \dots, 4. \tag{S4}$$

The corresponding virtual joint locations expressed in the distal plane are

$$\mathbf{C}_i = \begin{bmatrix} R_{Body}\cos\varphi_i \\ R_{Body}\sin\varphi_i \\ 0 \end{bmatrix}_{\Pi_E}.$$

The tail motor is located at the fixed point $\mathbf{T}$ with respect to the base-frame. The desired state consists of the distal-tip position $\boldsymbol{P}_{\boldsymbol{E}}^* \in \mathbb{R}^3$ and the arc-length coordinate $\boldsymbol{S}_{\boldsymbol{M}}^* \in \mathbb{R}$ of a tagged membrane point. The intermediate geometric span vector is

$$\mathbf{L} = [L_{Tail} \quad L_{W1} \quad L_{W2} \quad L_{W3} \quad L_{W4}]^T \in \mathbb{R}^5. \tag{S5}$$

Here, $L_{Tail}$ and $L_{Wi}$ denote geometric virtual-span lengths only. The actuator-side membrane travel is denoted by *d* below. The fixed half-torus length is excluded from this geometric-span vector and calculated as:

$$L_{Torus} = \pi R_{Torus}, \qquad R_{Torus} = \frac{R_{Body}}{2},$$

with $R_{Body} = 63.5$ mm. The actuator quadrature encoder state vector is

$$\mathbf{q} = [q_{Tail} \quad q_{W1} \quad q_{W2} \quad q_{W3} \quad q_{W4}]^T \in \mathbb{R}^5. \tag{S6}$$

Although the system has five actuators, the commanded state has four independent coordinates: the three components of $\mathbf{P}_E$ and the scalar $S_M$. The remaining actuator degree of freedom is constrained by membrane-length conservation.

**S5.2 Distal pose and virtual-string lengths**

The desired tip direction defines the distal-plane normal:

$$\boldsymbol{n}_{\boldsymbol{E}}^{*} = \frac{\boldsymbol{P}_{\boldsymbol{E}}^{*}}{\|\boldsymbol{P}_{\boldsymbol{E}}^{*}\|}, \qquad \mathbf{R}\mathbf{n}_0 = \boldsymbol{n}_{\boldsymbol{E}}^{*}. \tag{S7}$$

Thus, $\boldsymbol{n}_{\boldsymbol{E}}^{*}$ always points from $\mathbf{O}$ toward $\boldsymbol{P}_{\boldsymbol{E}}^{*}$, whereas $\mathbf{n}_0$ remains fixed in the base frame. Because rotation about this direction is excluded, $\boldsymbol{R}^{*}$ is the minimum rotation that maps $\mathbf{n}_0$ onto $\boldsymbol{n}_{\boldsymbol{E}}^{*}$. Let

$$\cos u = \mathbf{n}_0 \cdot \boldsymbol{n}_{\boldsymbol{E}}^{*}, \qquad \sin u = \| \mathbf{n}_0 \times \boldsymbol{n}_{\boldsymbol{E}}^{*} \|, \qquad \mathbf{a} = \frac{\mathbf{n}_0 \times \boldsymbol{n}_{\boldsymbol{E}}^{*}}{\|\mathbf{n}_0 \times \boldsymbol{n}_{\boldsymbol{E}}^{*}\|}, \tag{S8}$$

where $u$ and $\mathbf{a}$ are the rotation angle and unit rotation axis, respectively. The minimum rotation form of the Rodrigues' formula then gives

$$\boldsymbol{R}^{*} = \cos u\, \mathbf{I} + (1 - \cos u)\mathbf{a}\mathbf{a}^{T} + \sin u\, [\mathbf{a}]_{\times}, \tag{S9}$$

with

$$[\mathbf{a}]_{\times} = \begin{bmatrix} 0 & -a_z & a_y \\ a_z & 0 & -a_x \\ -a_y & a_x & 0 \end{bmatrix}. \tag{S10}$$

For the straight configuration ($u = 0$), the controller sets $\boldsymbol{R}^{*} = \mathbf{I}$. The antiparallel case is outside the commanded workspace because all desired normals lie in the forward hemisphere.

The center of the distal plane is offset from the most distal point of the half-torus by $R_{Torus}$:

$$\boldsymbol{G}^{*} = \boldsymbol{P}_{\boldsymbol{E}}^{*} - R_{Torus}\boldsymbol{n}_{\boldsymbol{E}}^{*}. \tag{S11}$$

Each wall string connects $\mathbf{B}_i$ to the transformed distal point $\boldsymbol{R}^{*}\mathbf{C}_i + \boldsymbol{G}^{*}$. Its vector and length are

$$\boldsymbol{l}_{\boldsymbol{Wi}}^{*} = \boldsymbol{R}^{*}\mathbf{C}_i + \boldsymbol{G}^{*} - \mathbf{B}_i, \qquad L_{Wi}^{*} = \| \boldsymbol{l}_{\boldsymbol{Wi}}^{*} \|, \qquad i = 1, \dots, 4. \tag{S12}$$

The tail string connects $\mathbf{T}$ to $\boldsymbol{G}^{*}$:

$$\boldsymbol{l}_{\boldsymbol{Tail}}^{*} = \boldsymbol{G}^{*} - \mathbf{T}, \qquad L_{Tail}^{*} = \| \boldsymbol{l}_{\boldsymbol{Tail}}^{*} \|. \tag{S13}$$

Equations (S7) to (S13) map the commanded tip position $\boldsymbol{P}_{\boldsymbol{E}}^{*}$ to the four wall and one tail geometric virtual-span lengths.

Relative to the current configuration, the required geometric span changes are:

$$\Delta L_{Tail} = L_{Tail}^{*} - L_{Tail,0}, \qquad \Delta L_{Wi} = L_{Wi}^{*} - L_{Wi,0}, \quad i = 1, \dots, 4. \tag{S14}$$

Under the taut, inextensible, and no-slip assumptions, the actuator-side membrane travel required to produce this geometric shape change equals the corresponding shape change:

$$\Delta d_{Tail}^{shape} = \Delta L_{Tail}, \qquad \Delta d_{Wi}^{shape} = \Delta L_{Wi}. \tag{S15}$$

### S5.3 Membrane motion and cycling correction

The tagged membrane point is described by the arc-length coordinate $S_M$ along a path comprising the tail, the half-torus, and the wall associated with the tool mounting angle $\varphi_{MP}$ (Fig. 3A(ii)). Substituting $\varphi_{MP}$ for $\varphi_i$ in Eq. (S4) defines the corresponding base and distal points $\mathbf{B}_{MP}$ and $\mathbf{C}_{MP}$. The transformed distal point and corresponding wall length are

$$\boldsymbol{D}^*_{\boldsymbol{MP}} = \boldsymbol{R}^*\mathbf{C}_{MP} + \boldsymbol{G}^*, \qquad L_{MP} = \| \boldsymbol{D}^*_{\boldsymbol{MP}} - \mathbf{B}_{MP} \|. \tag{S16}$$

The total path length is

$$L_{Total} = L_{Tail} + L_{Torus} + L_{MP}. \tag{S17}$$

Define the mean wall travel associated with the geometric shape change as

$$\Delta d^{shape}_{W\,Avg} = \frac{1}{4}\sum_{i=1}^{4} \Delta d^{shape}_{Wi}, \qquad \Delta S^{shape}_{M} = \frac{1}{2}\left(\Delta d^{shape}_{Tail} - \Delta d^{shape}_{W\,Avg}\right). \tag{S18}$$

Positive length changes denote payout. Equal tail and mean-wall payout produces no material motion relative to the tip; unequal payout everts or inverts the membrane through the tip. Equation (S18) is the supplementary form of Eq. (1) in the main text.

Let $S_O$ denote the arc-length coordinate of the most distal point of the half-torus

$$S^*_O = L^*_{Tail} + \frac{1}{2}L_{Torus}. \tag{S19}$$

Before the shape change, the tagged point may have an offset $\zeta_0 = S_{M,0} - S_{O,0}$. After the shape change, its predicted material coordinate $S_{M,pred}$ is

$$S_{M,pred} = S^*_0 + \zeta_0 + \Delta S^{shape}_{M}. \tag{S20}$$

The additional cycling displacement required to reach the commanded coordinate is

$$\Delta S_{Cycling} = S^*_M - S_{M,pred}. \tag{S21}$$

The final actuator commands combine the shape-change and cycling contributions

$$\Delta d^{cmd}_{Tail} = \Delta d^{shape}_{Tail} + \Delta S_{Cycling}, \tag{S22}$$

$$\Delta d^{cmd}_{Wi} = \Delta d^{shape}_{Wi} - \Delta S_{Cycling}, \qquad i = 1, \ldots, 4. \tag{S23}$$

The opposite signs circulate the membrane from the wall reservoir to the tail reservoir, or vice versa, while preserving the shape-change contribution to the distal pose. The final step converts these values into desired changes in quadrature encoder counts $\Delta\boldsymbol{q}$ using known parameters

$$\boldsymbol{q}^* = \boldsymbol{q}_0 + \Delta\boldsymbol{q} \tag{S24}$$

After each completed move, the commanded target is assigned as the current commanded state, $\boldsymbol{q}^* \rightarrow \boldsymbol{q}_0$

**S5.4 Cartesian reconstruction of the tagged membrane point**

For visualization and validation, $S_M$ is mapped to the Cartesian position $\mathbf{P}_M$ using separate expressions for the tail, torus, and wall regions. The coordinate increases from $S_M = 0$ at the tail motor to $S_M = L_{Total}$ at the selected wall roller.

For $0 \leq S_M \leq L_{Tail}$, let $\delta = S_M/L_{Tail}$. Then

$$\mathbf{P}_M(\delta) = \mathbf{T} + \delta(\mathbf{G} - \mathbf{T}). \tag{S25}$$

For $L_{Tail} < S_M \leq L_{Tail} + L_{Torus}$, let

$$\theta_{Torus} = \frac{S_M - L_{Tail}}{R_{Torus}} \in [0, \pi].$$

The tagged point follows the half-torus according to

$$\mathbf{P}_M(\theta_{Torus}) = \mathbf{C}_{Tip} + R_{Torus}[\cos\theta_{Torus}\,\mathbf{e}_1 + \sin\theta_{Torus}\,\mathbf{e}_2], \tag{S26}$$

where

$$\mathbf{C}_{Tip} = \mathbf{G} + \frac{1}{2}\mathbf{R}\mathbf{C}_{MP}, \qquad \mathbf{e}_1 = \frac{\mathbf{G} - \mathbf{C}_{Tip}}{\|\mathbf{G} - \mathbf{C}_{Tip}\|}, \qquad \mathbf{e}_2 = \mathbf{n}_E. \tag{S27}$$

For $L_{Tail} + L_{Torus} < S_M \leq L_{Total}$, let

$$\tau = \frac{S_M - L_{Tail} - L_{Torus}}{L_{MP}} \in [0,1].$$

The tagged point then moves from the distal endpoint $\mathbf{D}_{MP}$ toward the wall roller $\mathbf{B}_{MP}$:

$$\mathbf{P}_M(\tau) = (1 - \tau)\mathbf{D}_{MP} + \tau\mathbf{B}_{MP}. \tag{S28}$$

These piecewise expressions are continuous at the tail–torus and torus–wall boundaries.

**S5.5 Hardware implementation**

A Teensy 4.1 microcontroller coordinated the five Maxon EPOS2 drives over a 1 Mbit/s CAN bus. The four wall rollers and tail spool were triggered by a common synchronization signal, and their velocities were scaled so that all commanded length changes ended simultaneously. Low level position and velocity control was managed by individual EPOS2 controllers. The encoders provided 1024 counts per revolution in quadrature (× 4), and each actuator used a 190:1 gear reduction. The calibrated effective diameters were 36.6 mm for the tail spool and 44.32 mm for the wall rollers.

The higher-level supervisory controller supported shape-only motion, membrane cycling, and combined shape change and cycling compensation commands. After a quick-stop event, the commanded state was reset to the measured encoder positions before motion resumed. Body pressure and the two finger pressures were regulated independently through three pulse-width-modulated channels updated at 50 Hz.

Table S1. Kinematics nomenclature

| Symbol | Definition |
|---|---|
| $\boldsymbol{P}_E$ | Distal tip position |
| $\boldsymbol{P}_E^*$ | Commanded distal tip position |
| $S_M$ | Scalar membrane point coordinate |
| $S_M^*$ | Desired membrane point coordinate |
| $S_O$ | Membrane point reference (always at most distal point of torus) |
| $S_O^*$ | Latest membrane point reference |
| $S_{O,0}$ | Initial membrane point reference |
| $\zeta_0$ | Initial membrane point offset from reference |
| $\mathbf{q}$ | Actuator command vector |
| $\boldsymbol{q}^*$ | Desired actuator command vector |
| $\Delta S_{Cycling}$ | Additional membrane displacement required to reach $S_M^*$ |
| $\mathbf{n}_0$ | Base frame platform normal |
| $\mathbf{n}_E$ | Distal frame platform normal |
| $\boldsymbol{R}$ | Rotation matrix mapping vectors from the distal frame to the base frame |
| $\Delta d_{Wi}^{shape}, \Delta d_{Tail}^{shape}$ | Actuator-side membrane travel required for the geometric shape change |
| $\Delta d_{Wi}^{cmd}, \Delta d_{Tail}^{cmd}$ | Final actuator-side membrane travel after cycling correction |
| $\boldsymbol{G}$ | Distal platform center position |
| $\boldsymbol{O}$ | Base platform origin |
| $\varphi_i$ | Angle describing the location of virtual joints on their respective planes |
| $\boldsymbol{B}_i$ | Base virtual joint of wall string |
| $\boldsymbol{C}_i$ | Corresponding distal-platform anchor expressed in $\Pi^E$ |

## S6. Kinematic validation protocol

Each trajectory was validated using two 6-DoF electromagnetic micro sensors (trakSTAR, Northern Digital) attached to the deployed membrane. The sensor at the distal point of the torus tracked the membrane material point, $\mathbf{P_M}$, whereas the sensor at the torus center tracked the end-effector position, $\mathbf{P_E}$. To guarantee stable measurements, each sensor was inserted inside of a 3D printed polymer fixture that was attached to the membrane using double sided tape (for $\mathbf{P_M}$ ) or with a rod concentrically fastened to the tail with double sided tape (for $\mathbf{P_E}$). Measurements were acquired at 80 Hz and transformed from the transmitter frame to the robot base frame using the known pose of a laser-cut mounting fixture.

Before each trial, the robot was inflated to 4 kPa and fully extended to establish the initial body length and membrane coordinates. It was then retracted to a common starting configuration, and each trajectory was repeated five times. Planar and axial trajectories were evaluated using $\mathbf{P_E}$, projected onto the corresponding plotting plane, whereas membrane cycling was evaluated using $\mathbf{P_M}$ because the end-effector pose remained constant. Absolute waypoint error was calculated as the Euclidean distance between the commanded and measured positions after settling. Results are reported as the mean ± SD across five trials. The experimental setup used for kinematic validation is shown in Fig. S5.

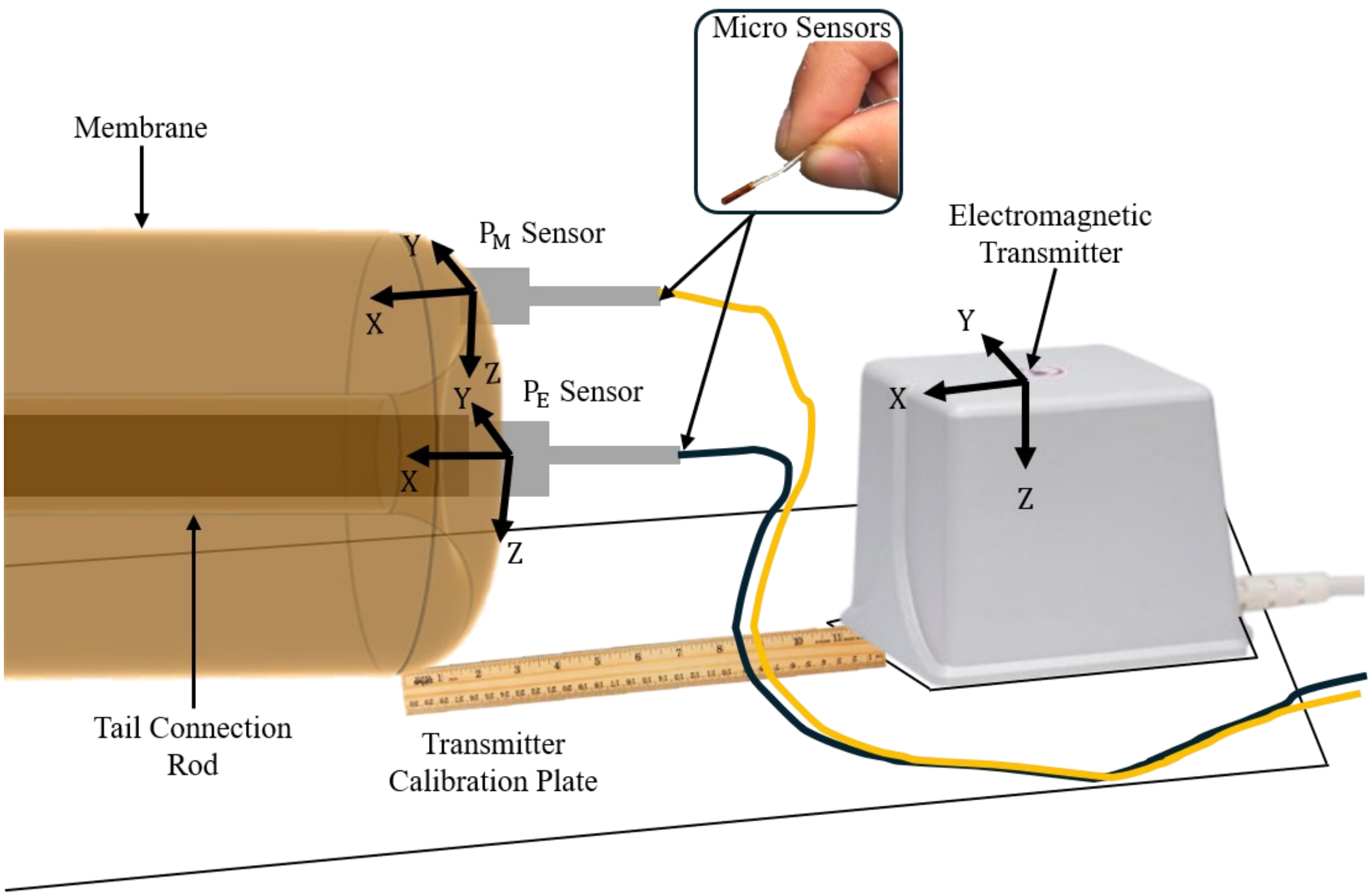

**Fig. S5. Kinematic validation setup.** Two 6-DoF electromagnetic sensors are mounted on the deployed membrane and read out by a trakSTAR transmitter that is seated in a laser cut acrylic fixture of known pose relative to the robot base. The probe for $P_M$ is bonded to the most distal point of the torus with double sided tape and is used to evaluate the membrane cycling motions, where material transport occurs at constant end-effector pose. The $P_E$ sensor is attached to the geometric center of the torus aligned coplanar with the distal torus point with a rod and is used to evaluate all pose-changing trajectories.

## S7. Pleated finger design and fabrication protocol

The fabrication workflow of the pleated inflatable finger is shown in Fig. S6. The textile shell was fabricated from TPU-coated nylon using a flat pattern that included the material required for the dorsal pleats and an 11-mm seam allowance. Pleat fold lines were lightly engraved using a VersaLaser VLS3.50 $CO_2$ laser cutter at minimum power and maximum speed, without cutting through the fabric. The dorsal layer was folded along the engraved lines, temporarily secured with clips at the prescribed pleat spacing, and stitched to the smooth ventral layer. Cotton thread approximately 0.25 mm in diameter was used, with a stitch pitch of approximately 0.75 mm. The lateral edges were progressively closed while leaving a 30-mm opening at the proximal end for bladder insertion and attachment to the integration interface. The proximal end was pinned and stitched to a circular textile corolla with an internal diameter of 40 mm and an external diameter of 60 mm. The finger–corolla assembly was then attached to a TPU-coated circular layer, leaving an exposed TPU border for thermal sealing to the TPU-coated robot body. An oversized 0.05-mm-thick TPU bladder was cut using the $CO_2$ laser in vector mode at 20% power and 80% speed and inserted through the proximal opening. After connection of the pneumatic tube, the completed finger was inflated and inspected to confirm proper pleat deployment, symmetric bending, and airtightness.

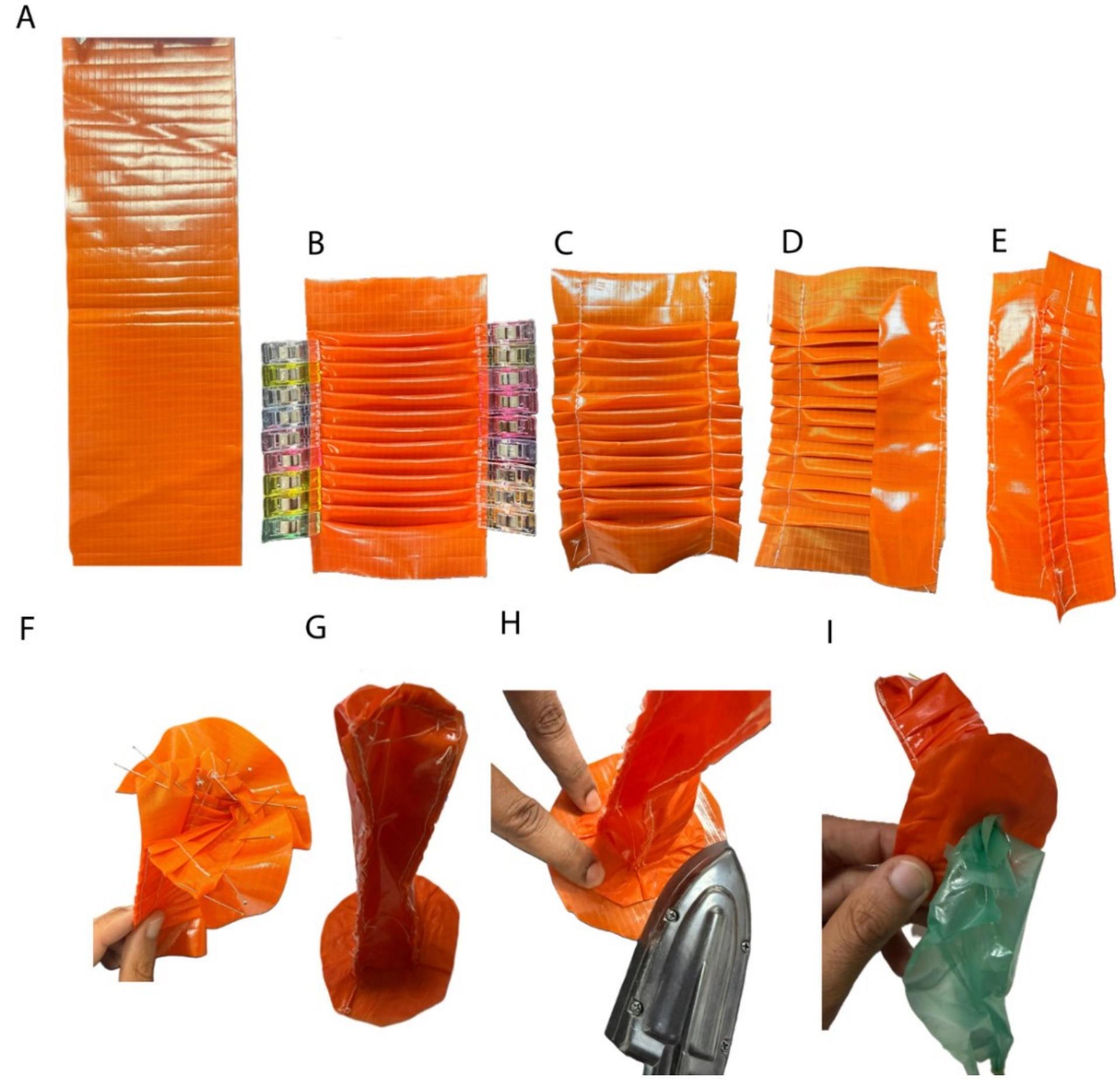

**Fig. S6. Fabrication workflow of the pleated inflatable finger and circular integration interface**. (A) TPU-coated nylon flat pattern with engraved fold lines. (B) Pleating and temporary fixation using clips. (C) Stitching of the pleated and smooth layers. (D and E) Progressive closure of the lateral edges, leaving a proximal opening. (F and G) Attachment of the finger to the circular textile corolla. (H) Addition of a TPU-coated circular layer for thermal sealing to the robot body. (I) Insertion of the internal TPU bladder.

## S8. Integration of fingers on the everting membrane

After fabrication, the pleated-finger module was integrated into the TPU-coated nylon membrane of the soft everting robot (Fig. S7). The module was inserted from inside the body through a local opening, with its circular TPU-coated border aligned against the inner membrane surface (Fig. S7A). The border was thermally bonded to the body using a handheld heat-sealing tool (Clover Mini Iron II) at approximately 200°C. Manual pressure was applied for 10–15 s through a PTFE release film, followed by cooling under pressure for 20–30 s (Fig. S7B). Heating was restricted to the interface perimeter to avoid damaging the pleated region or internal bladder. The open edge of the robot body was subsequently closed using a PFS-300 impulse sealer with an impulse time of 0.8–1.2 s and a 3–5 s cooling period. The completed assembly (Fig. S7C) was inspected for continuous bonding and tested at low pressure to identify leakage, delamination, or obstruction of finger motion.

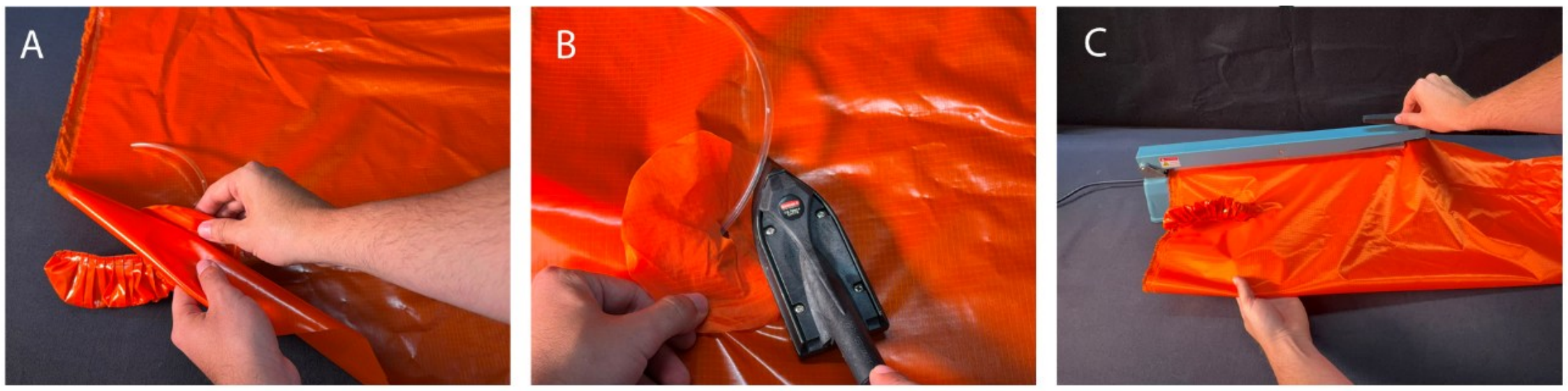

**Fig. S7. Integration of the pleated finger module into the soft everting body.** **(A)** Positioning of the finger module inside the TPU-coated nylon membrane of the soft everting robot. The circular interface is aligned with the selected opening in the body. **(B)** Thermal sealing of the TPU-coated circular border to the TPU-coated vine membrane using a handheld heat-sealing tool. The seal is applied around the perimeter while avoiding the active pleated region and internal bladder. **(C)** Final integrated membrane after thermal sealing with impulse sealer and flattening, showing the finger module embedded in the vine body before inflation and functional testing.

## S9. Finger bending characterization

To characterize pressure-dependent bending, each finger was clamped at its proximal base and photographed orthogonally to the bending plane. A coplanar calibration marker was used to convert image coordinates into physical dimensions. The fingers were inflated from 0 to 50 kPa in 10-kPa increments and held for 10 s before image acquisition. High-contrast centerline markers were used to track deformation, with 11 markers on the 120-mm finger and 7 on the 70-mm finger. Marker coordinates were manually extracted and fitted with a cubic smoothing spline to reconstruct the finger centerline. Under a constant-curvature assumption, a circular arc was fitted to the centerline, and the bending angle was calculated as $\theta = \kappa L$, where $\kappa$ is the fitted curvature and $L$ is the undeformed finger length.

## S10. Blocking-force characterization

Blocking-force tests were performed using a digital force gauge (SAUTER FK 10, 10 N capacity) mounted on a rigid support to quantify the force generated by the finger when its bending motion was mechanically constrained. Each finger was clamped at its proximal base in the same fixture used for the bending characterization, reproducing the boundary condition at the robot body. The distal region of the finger was positioned in contact with the flat probe of the force gauge. The probe was aligned approximately normal to the expected direction of distal-tip motion, so that the measured force corresponded to the blocked bending force generated during pressurization. Before each trial, the force gauge was zeroed and the finger was positioned so that the distal tip was just in contact with the probe without measurable preload. The finger was then inflated from 0 to 50 kPa in 10 kPa increments. At each pressure level, the pressure was held for 10 s to allow the pneumatic chamber and fabric structure to reach a quasi-static condition. The blocking force was recorded after stabilization, and the steady-state value was used for analysis. The test was repeated for both finger lengths, $L = 120$ mm and $L = 70$ mm, using the same clamping and contact configuration.

## S11. Workspace-map construction

Workspace maps were generated from the experimentally measured finger centerlines at pressures from 0 to 50 kPa in 10-kPa increments. The coordinate origin was placed midway between the two toroidal base-circle centers, with X in the lateral direction and Y in the distal direction. Each measured centerline was positioned on a toroidal base of radius 31.75 mm with a fixed inward angle offset, and the opposing finger was generated by mirror symmetry. For each pressure, the finger pair was swept over the prescribed toroid-angle range, and the polygons bounded by the finger centerlines and their corresponding base arcs were unioned to obtain the swept workspace.

Enclosure regions were obtained by retaining configurations in which the distal finger orientation indicated inward closure. The distal tangent angle was estimated from a cubic fit to the final four centerline points. For each retained configuration, the regions bounded by the left and right finger centerlines and the toroidal bases were combined across the angular sweep. To reduce visual overlap, only the 10-, 30-, and 50-kPa enclosure regions are shown. This pressure subsampling was used only for visualization and did not define a different enclosure criterion.

Object-size feasibility was evaluated separately for circular cross-sections centered on the gripper midline. For each measured pressure and toroid angle, the largest cylinder satisfying the finger-clearance, fingertip-height, and centerline-crossing constraints was determined using geometric radius bounds followed by binary search. The resulting maximum diameter was plotted as a function of pressure and toroid angle.

The fingertip-force panel provides a directional geometric analysis rather than a prediction of grasp-force magnitude. A unit force normal to the distal finger tangent was decomposed into normal and tangential components at representative toroid angles. Pull-out force and stiffness were measured experimentally and are reported separately in the main text. Thus, these maps provide geometric and quasi-static estimates of workspace, enclosure, and object-size compatibility which may be used with experimental results to provide dynamic retention and grasp quality estimates. The physical characteristics of the objects used in the grasping experiments are summarized in table S2.

Table S2. Physical characteristics of the objects used in the grasping experiments

| Object | Weight (g) | Main dimension A (mm) |
|---|---|---|
| Plastic screw | 92 | 16 |
| Screwdriver | 98 | 36 |
| Domino box | 453 | 37 |
| Strawberry | 21 | 50 |
| Rectangular object | 59 | 58 |
| Cylindrical object | 55 | 64 |
| Blue mug | 60 | 72 |
| Ceramic mug | 550 | 82 |
| Ball | 100 | 200 |

## S12. Distal body stiffness characterization in the fingertip configuration.

Distal stiffness was characterized only in the fingertip configuration. In this test, the variable parameter was the pressure of the vine body, not the finger actuation pressure. The gripper body was fixed to a rigid frame and inflated to body pressures of 5, 10, 20, 30, and 40 kPa. The finger pressure was kept constant during the tests at 50kPa, so that changes in measured stiffness could be attributed to the pressurization state of the vine body. A six-axis force/torque sensor (ATI Nano17, ATI Industrial Automation, USA) was mounted on a motorized linear translation stage and contacted the distal finger region with a cylindrical contact surface covered by a high friction tape (3M TB641) to minimize slip. The stage was translated at 1 mm/s over a 20 mm displacement range, while the force component along the pushing direction was recorded at 500 Hz. For each body-pressure level, the system was held for 10 s before testing to allow the vine body and integrated finger structure to reach a quasi-static configuration. The force signal was baseline-corrected using the first 0.5 s of data. The force trace was then aligned by identifying the maximum force, which corresponded to the end of the imposed pulling motion. A 20 s window preceding this maximum force was extracted for analysis, corresponding to 20 mm of stage displacement at the imposed speed of 1 mm/s. For each trial, fingertip stiffness was computed as the slope of a linear fit to the force–displacement response over this 20 mm displacement window. Since the stage speed was 1 mm/s, the slope obtained from force versus time in N/s was equivalent to stiffness in N/mm. Up to five trials were analyzed for each body pressure, and the reported stiffness corresponds to the mean of the per-trial linear fits, with variability reported across trials. The experimental setup used to characterize distal stiffness is shown in Fig. S8.

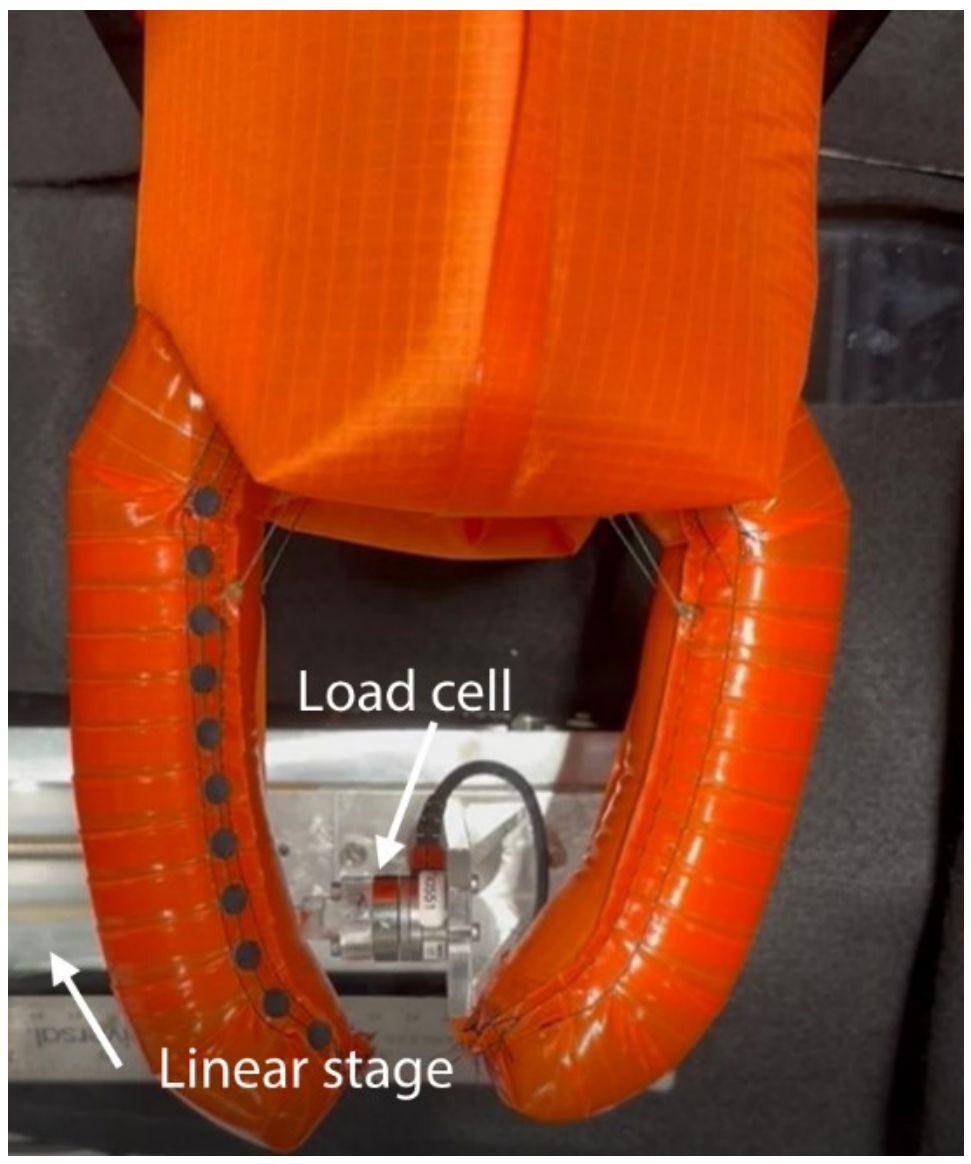

**Fig. S8. Experimental setup for distal-stiffness characterization.** The distal stiffness of the finger–body module was measured in the fingertip configuration. The vine body was fixed in a vertical configuration and pressurized to the target body pressure. The distal region of one finger was connected through a thin cable to an ATI Nano17 force/torque sensor mounted on a motorized linear stage. The stage imposed a controlled indentation displacement, while the force along the pulling direction was recorded and used to compute the effective fingertip stiffness.

## S13. Data processing and statistical analysis

All data reported in the Supplementary Information were processed using custom MATLAB scripts. Image-derived measurements were converted from pixels to physical units using calibration markers or reference dimensions included in the images. Landmark or marker coordinates were used when needed to reconstruct soft-structure geometries, extract centerlines, estimate bending angles, or generate workspace maps. Force data were zeroed or baseline-corrected before analysis, and the force component aligned with the loading direction was used. Depending on the experiment, the reported quantity corresponded to the stabilized force, the maximum force before slip or release, or the slope of the force–displacement response. For stiffness measurements, stiffness was computed from the linear region of the force–displacement curve over the prescribed displacement range. For pressure-dependent tests, data were grouped by actuation condition and repeated trials were analyzed for each pressure level. Continuous variables are reported as mean ± standard deviation unless otherwise stated. Line plots show mean values with shaded regions indicating variability, while box plots summarize the distribution across repeated trials. Single-trial functional demonstrations and object-retention tests were reported descriptively and were not used to estimate success rates. The analyses were descriptive and were used to compare trends across configurations and operating conditions.